\documentclass{article}

\usepackage{geometry}
\usepackage{natbib}%
\usepackage[figuresright]{rotating}
\usepackage{xcolor}

\usepackage{dutchcal}
\usepackage{authblk}

\usepackage{url}
\usepackage{enumitem} 

\usepackage{algorithmic}
\usepackage{algorithm}
\usepackage{multirow}

\usepackage{amsfonts}
\usepackage{mathtools}

\DeclarePairedDelimiter\floor{\lfloor}{\rfloor}

\title{Optimal Decision Trees for the Algorithm Selection Problem: Integer Programming Based Approaches}

\author[1]{Matheus Guedes Vilas Boas \thanks{matheusgueedes91@gmail.com}}
\author[1,2]{Haroldo Gambini Santos \thanks{haroldo@ufop.edu.br}}
\author[3]{Luiz Henrique de Campos Merschmann\thanks{luiz.hcm@dcc.ufla.br}} 
\author[2]{Greet Vanden Berghe\thanks{greet.vanden.berghe@kuleuven.be }}

\affil[1]{Department of Computing, Federal University of Ouro Pretoo, 
St. Diogo de Vasconcelos 328, 35400-000, Ouro Preto, MG, Brazil}
\affil[2]{Computer Science Technology Campus, University of KU Leuven, 9000, Ghent, Belgium}
\affil[3]{Department of Computer Science, Federal University of Lavras, P.O.Box 3037, 
37200-000, Lavras, MG, Brazil}

\begin{document}

\maketitle

\begin{abstract}
Even though it is well known that for most relevant computational problems different algorithms may perform better on different classes of problem instances, most researchers still focus on determining a single best algorithmic configuration based on aggregate results such as the average. In this paper, we propose Integer Programming based approaches to build decision trees for the Algorithm Selection Problem. These techniques allow automate three crucial decisions: ($i$) discerning the most important problem features to determine problem classes; ($ii$) grouping the problems into classes and ($iii$) select  the best algorithm configuration for each class. To evaluate this new approach, extensive computational experiments were executed using the linear programming algorithms implemented in the COIN-OR Branch \& Cut solver across a comprehensive set of instances, including all MIPLIB benchmark instances. The results exceeded our expectations. While selecting the single best parameter setting across all instances decreased the total running time by 22\%, our approach decreased the total running time by 40\% on average across 10-fold cross validation experiments. These results indicate that our method generalizes quite well and does not overfit.
\end{abstract}

\section{Introduction}\label{sec:Introduction}

Given that different algorithms may perform better on different classes of problems, \cite{RICE197665} proposed a formal definition of the Algorithm Selection Problem (ASP).  The main components of this problem are depicted in Fig. \ref{fig:rice}. Formally the ASP has the following input data:

\begin{figure}
\begin{center}
\includegraphics[]{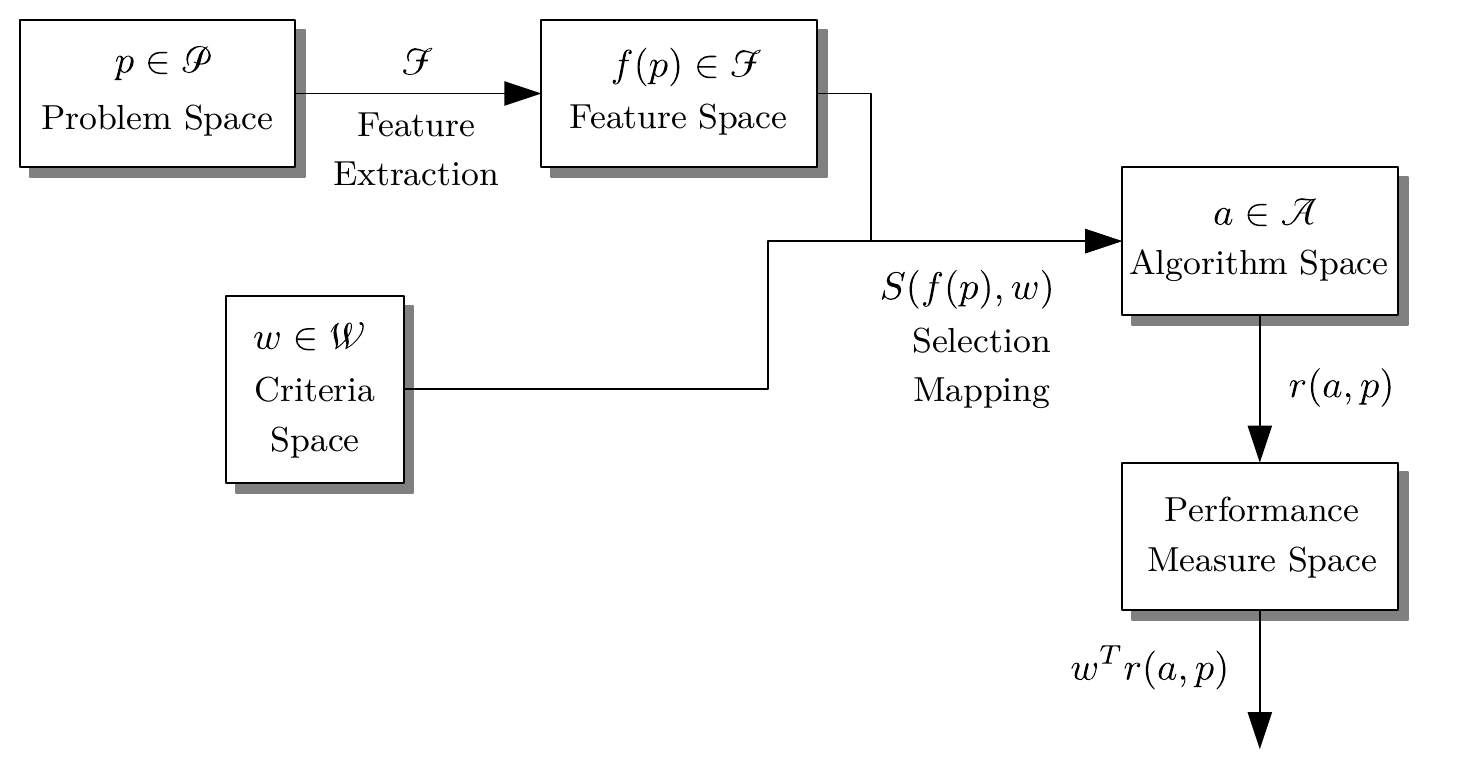}
\label{fig:rice} \caption{The Algorithm Selection Problem \citep{RICE197665} }
\end{center}
\end{figure}

\begin{description}

\item[{$\mathcal{P}$}]: the problem space, a probably very large and diverse set of different problem instances; these instances have a number of characteristics, i.e. for linear programming, each possible constraint matrix defines a different problem instance;

\item[{$\mathcal{A}$}]: the algorithm space, the set of available algorithms to solve instances of problem $\mathcal{P}$; since many algorithms have parameters that significantly change their behavior, differently from \cite{RICE197665}, we consider that each element in $\mathcal{A}$ is an algorithm with a specific parameter setting; thus, selecting the best algorithm also involves selecting the best parameter tuning for this algorithm;

\item[{$\mathcal{F}$}]: the feature space; ideally elements of $\mathcal{F}$ have a significantly lower dimension than elements of $\mathcal{P}$, since not every problem instance influences the selection of the best algorithm; these features are also important to cluster problems having a common best algorithm, e.g., for linear programming some algorithms are known for performing well on problems with a dense constraint matrix;

\item[{$\mathcal{W}$}]: the criteria space, since algorithms can be evaluated with different  criteria, such as processing time, memory consumption and simplicity, the evaluation of the execution results $r=\mathbb{R}^n$ produced using an algorithm $a$ to solve a problem instance $p$ may be computed using a weight vector $w \in \mathcal{W} = [0,1]^n$ which describes the relative importance of each criterion.

\end{description}

The objective is to define a function $S$ that, considering problem features, maps problem instances to the best performing algorithms. This function is a mapping function that always selects the best algorithm for every instance. Thus, if $B$ is the ideal function, the objective is to define $S$ minimizing:

\begin{equation}
 \sum_{p \in \mathcal{P}} | w^T r(B(f(p), w)) - w^T r(S(f(p), w))| 
\label{eqOFASP}
\end{equation}

The main motivation for solving the ASP is that usually there is no \emph{best algorithm} in the general sense: even though some algorithms may perform better on average, usually some algorithms perform much better than others for some groups of instances. A ``winner-take-all'' approach will probably discard algorithms that perform poorly on average, even if they produce excellent results for a small, but still relevant, group of instances.

This paper investigates the construction of $S$ using decision trees. The use of decision trees to compute $S$ was one of the suggestions included in the seminal paper of \cite{RICE197665}. To the best of our knowledge however, the use of decision trees for algorithm selection was mostly ignored in the literature. One recent exception is the work of \cite{polyakovskiy2014comprehensive} who evaluated many heuristics for the traveling thief problem and built a decision tree for algorithm recommendation. \cite{polyakovskiy2014comprehensive} did not report which algorithm was used to build this tree, but did note that the MatLab\textsuperscript{\tiny\textregistered}  Statistics Toolbox was used to produce an initial tree that was subsequently pruned to produce a compact tree. This is an important consideration: even though deep decision trees can achieve 100\% of accuracy in the training dataset, they usually overfit, achieving low accuracy when predicting the class of new instances. The production of compact and accurate decision trees is an NP-Hard problem \citep{HYAFIL197615}. Thus, many greedy heuristics have been proposed, such as ID3 \citep{Quinlan:1986:IDT:637962.637969}, C4.5 \citep{Quinlan:1993:CPM:152181} and CART \citep{breiman}. These heuristics recursively analyze each split in isolation and proceed recursively.
Recently, \citep{Bertsimas:2017:OCT:3123655.3123731} proposed Integer Programming  for producing \emph{optimal} decision trees for classification. Thus, the entire decision tree is evaluated to reach global optimality. Their results showed that much better classification trees were produced for an extensive test set. This result was somewhat unexpected since there is a popular belief that optimum decision trees could overfit at the expense of generalization. Trees are not only the organizational basis of many machine learning methods, but also an important structural information \citep{zhang2018tree2vector}. The main advantage of methods that produce a tree as result is the  interpretability of the produced model, an important feature in some applications such as healthcare.

At this point it is important to clarify the relationship between decision trees for the ASP and decision trees for \emph{classification} and \emph{regression},  their most common applications. Although the ASP can be seen as the classification problem of selecting the best algorithm for each instance, this modeling does not capture some important problem aspects. Firstly, it is often the case that many algorithms may produce equivalent results for a given instance, a complication which can be remedied by using multi-label classification algorithms \citep{tsoumakas2007multi}. Secondly, the evaluation of the individual decisions of a classification algorithm always returns zero or one for incorrect and correct predictions, respectively. In the ASP each decision is evaluated according to a real number which indicates \emph{how far} the performance of the suggested algorithm is from the performance of the best algorithm, as stated in the objective function (\ref{eqOFASP}). Thus, the construction of optimal decision trees for the ASP can be seen as a generalization of the multi-label classification problem and is at least as hard. Another  approach is to model the ASP as a regression problem, in which one attempts to predict the performance of each algorithm for a given instance and select the best one \citep{xu2008satzilla,leyton2003portfolio,battistutta2017feature}. In this approach, the cost of determining the recommended algorithm for a new instance grows proportionally to the number of available algorithms and the performance of the classification algorithm. By contrast, a decision tree with limited depth can recommend an algorithm in constant time.  Another shortcoming of regression-based approaches is related to the loss of precision in solution evaluation:  consider the case when the results produced by a regression algorithm are always the correct result \emph{plus} some additional large constant value. Even though the ranking of the algorithms for a given instance would remain correct and the right algorithm would always be selected, this large constant would imply an (invalid) estimated error for the regression algorithm. The present paper therefore investigates the applicability of Integer Programming to build a mapping $S$ with decision trees using the precise objective function (\ref{eqOFASP}) of the ASP.

It is also important to distinguish the ASP from the problem of discovering improved parameter settings. Popular software packages such as irace \citep{irace} and ParamILS \citep{DBLP:journals/corr/HutterSLH14} embed several heuristics to guide the search of improved parameters to a parameter setting that, ideally, performs well across a large set of instances. During this search, parameter settings with a poor performance in an initial sampling of instances may be discarded. In the ASP, even parameter settings with poor results for many instances may be worth investigating since they may well be the best choice to a small group of instances with similar characteristics. Thus, exploring parameter settings for the ASP may be significantly more computationally expensive than finding the best parameter setting on average, requiring more exhaustive computational experiments. An intermediate approach was proposed by \cite{Kadioglu:2010:IIA:1860967.1861114}: initially the instance set is divided into clusters and then the parameter settings search begins. One shortcoming of this approach is the requirement of an \emph{a priori} distance metric to cluster instances. It can be hard to decide which instance features may be more influential for the parameter setting phase before the results of an initial batch of experiments is available. Optimized decision trees for the ASP provide important information regarding which instance features are more influential to parameter settings since these parameters will appear in the first levels of the decision tree. Also, instances are automatically clustered in the leaves. It is important to observe that an iterative approach is possible: after instances are clustered using a decision tree for the ASP, a parallel search for better parameters  for instance groups may be executed, generating a new input for the ASP.

Another fundamental consideration is that the ASP is a \emph{static} tuning technique: no runtime information is considered to dynamically change some of the suggested algorithm/parameter settings, as in the so called reactive methods \citep{mascia2014an}. The static approach has the advantage that usually no considerable additional computational effort is required to retrieve a recommended setting, but its success obviously depends on the construction of a sufficiently diverse set of problem instances for the training dataset to cover all relevant cases. After the assembly of this dataset, a possibly large set of experiments must be performed to collect the results of many different algorithmic approaches for each instance. Finally, a clever recommendation algorithm must be trained to determine relevant features for recommending the best parameters for new problem instances. \cite{MISIR2017291} tackle the problem of recommending algorithms with incomplete data, i.e., if the experiments results matrix is sparse and only a few algorithms were executed for each problem instance. In this paper we consider the more computationally expensive case, where for the training dataset all problem instances were evaluated on all algorithms.

This paper proposes the construction of optimal decision trees for the ASP using Integer Programming techniques. To accelerate the production of high quality feasible solutions, a variable neighborhood descent based mathematical programming heuristic was also developed. To validate our proposal, we set ourselves the challenging task of improving the performance of the COIN-OR Linear Programming Solver - CLP, which is the Linear Programming (LP) solver employed within the COIN-OR Branch \& Cut - CBC solver \citep{Lougee-Heimer2003}. CLP is currently considered  the \emph{fastest} open source LP solver \citep{hansmtbenchlp,Gearhart13}. The LP solver is the main component in Integer Programming solvers \citep{atamturk2005integer} and it is executed at every node of the search tree. Mixed-Integer Programing is the most successful technique to optimally solve NP-Hard problems and has been applied to a large number of problems, from production planning \citep{Pochet:2006:PPM:1202598} to prediction of protein structures \citep{zhu}.

To the best of our knowledge, this is the first time that mathematical programming based methods have been proposed and computationally evaluated for the ASP. As our results demonstrate, not only our algorithm produces more accurate predictions for the best algorithm with respect to unknown instances, considering a 10-fold validation process (Section 4.3) but it also has the distinct advantages of recommending algorithms in constant time and producing easily interpretable results.

The remainder of the paper is organized as follows: Section \ref{sec:Model} presents the Integer Programming formulation for the construction of optimal decision trees for the ASP.
Section \ref{sec:vnd} presents a variable neighborhood descent based mathematical programming heuristic. Our extensive computational experiments and their results are presented in Section \ref{sec:Experiments} and, finally, Section \ref{sec:Conclusions} presents the results and provides some future research directions.

\section{Optimal Decision Trees for the Algorithm Selection Problem}\label{sec:Model}
This section presents our integer programming model proposed for the construction of optimal decision trees for the ASP. The sets and parameters are described in Section \ref{inputData}. The corresponding decision variables are described in Section \ref{decisionVariables}. The objective function associated with the problem and the constraints are described in Section \ref{foandconstraints}.

\subsection{Input data}\label{inputData}

\begin{description}[leftmargin=!,labelwidth=\widthof{\bfseries label}]

\item[{$\mathcal{P}$}] set of problem instances = $\{1,\ldots,\overline{p}\}$;
\item[{$\mathcal{A}$}] set of available algorithms with parameter settings = $\{1,\ldots,\overline{a}\}$;
\item[{$\mathcal{F}$}] set of instance features = $\{1,\ldots,\overline{f}\}$;
\item[{$\mathcal{C_f}$}] set of valid branching values for feature $f$, $C_f=\{1,\ldots,\overline{c}_f\}$, ${\overline{c}}_f$ is at most $\overline{p}$ when all instances have different values for feature $f$;
\item[$d$] maximum tree depth;
\item[$\tau$] threshold indicating a minimum number of instances per leaf node;
  \item[$\beta$] penalty incorporated into the objective function when a leaf node contains a number of problem instances smaller than threshold $\tau$;
\item[$v_{p,f}$] value of feature $f$ for  problem instance $p$;
\item[$g_{l,n}$] parameter that indicates which is the parent node of a given node $n$ (considering a child node $n$ where its parent is at its left);
\item[$h_{l,n}$] parameter that indicates which is the parent node of a given node $n$ (considering a child node $n$ where its parent is at its right);

\end{description}

 The maximum allowed tree depth is defined by $d$. To prevent overfitting, an additional cost (parameter $\beta$)  is included into the objective function to penalize the occurrence of leaf nodes containing a number of problem instances smaller than threshold $\tau$.

Parameters $v_{p, f}$ indicate the value of each feature $f$ for each problem instance $p$. Parameters $g_{l,n}$ and $h_{l, n}$ indicate the parent node of a given node located at the left or right, respectively. Thus, if the parent of the node $n$ is at left ($n \mod 2 = 0$), then $g_{l,n} = \floor*{(n+1)/2}$, otherwise $g_{l,n} = -1$. Similarly, if the parent of the node $n$ is at the right ($n \mod 2 = 1$), then $h_{l,n} = \floor*{(n+1)/2}$, otherwise, $h_{l,n} = -1$.

\subsection{Decision variables}\label{decisionVariables}

The main decision variables $x_{l,n,f,c}$ are related to the feature and branch values at each branching node of the decision tree. From the choices defined by the model, the problem instances are grouped (variables $y_{l, n, p}$) according to the features and the cut-off points that were imposed on the branches. The model will determine the best algorithm for each group of problem instances placed on each leaf node (variables $z_{n, a}$). Variables $u_n$ are used to check if there are problem instances allocated to a given leaf node. This set will be linked to the set of variables $m_n$ - explained later in this section.

\begin{align*}
x_{l,n,f,c} ={}& \left\{\begin{aligned}
&1, \text{if feature $f \in \mathcal{F}$ and cut-off point $c \in \mathcal{C}_f$ is used for node $n \in \{1,\ldots,2^{l}\}$} \\ & \qquad \text{ of level $l \in \{0,\ldots,(d-1)\}$.}\\
&0, \text{otherwise}.
\end{aligned}\right.\\\\
y_{l,n,p} ={}& \left\{\begin{aligned}
&1, \text{if problem instance $p \in \mathcal{P}$ is included for node $n \in \{1,\ldots,2^{l}\}$} \\ & \qquad \text{of level $l \in \{1,\ldots,d\}$.}\\
&0, \text{otherwise}.
\end{aligned}\right.\\\\
z_{n,a} ={}& \left\{\begin{aligned}
&1, \text{if algorithm $a \in \mathcal{A}$ is used in the leaf node $n \in \{1,\ldots,2^{d}\}$.}\\
&0, \text{otherwise}.
\end{aligned}\right.\\\\
u_{n} ={}& \left\{\begin{aligned}
&1, \text{if leaf node $n \in \{1,\ldots,2^{d}\}$ has problem instances.}\\
&0, \text{otherwise}.
\end{aligned}\right.
\end{align*}

The next two sets of decision variables are used in the objective function. With the exception of set $m_{n}$ ($m_{n} \in \mathbb{Z}^{+}$), all other sets of variables are binary. To penalize leaf nodes with few instances, which could result in overfitting, variables $m_n$ are used to compute the number of problem instances that are missing for the leaf node $n$ to reach a pre-established threshold of problem instances per leaf node, determined by the parameter $\tau$. The set of decision variables $w_{p, n, a}$ is responsible for connecting the sets of decision variables $y_{l, n, p}$ and $z_{n, a}$, i.e., to ensure that all problem instances allocated to a particular leaf node have the same recommended algorithm and that this algorithm is exactly the one corresponding to $z_{n, a}$. In addition, the connection between the set $w_{p, n, a}$ and $y_{l, n, p}$ ensures that the problem instances allocated to leaf nodes respect branching decisions on parent nodes.

\vspace{0.15in}

\begin{align*}
m_{n} ={}& \left\{\begin{aligned}
& \text{number of problem instances missing from the leaf node $n$ to reach}\\  & \qquad{\text{a pre-established threshold of problem instances per leaf node.}}\\
\end{aligned}\right.\\\\
w_{p, n, a} ={}& \left\{\begin{aligned}
&1, \text{if problem instance $p \in \mathcal{P}$ is selected for leaf node $n \in \{1,\ldots,2^{d}\}$} \\ & \qquad{\text{with algorithm $a \in \mathcal{A}$.}}\\
&0, \text{otherwise}.
\end{aligned}\right.\\
\end{align*}

\subsection{Objective function and constraints}\label{foandconstraints}
The objective of our model is to construct a tree of determined maximum depth that minimizes the  distance of the performance obtained using the recommended algorithm from the ideal performance for each problem $p$. Here we consider that this non-negative value is already computed in $r$. There is an additional cost involved in the objective function to penalize the occurrence of leaf nodes with only a few supporting instances. Follows the objective function (\ref{eq:1}) and the set of constraints (\ref{eq:2}-\ref{eq:20}) of our model:

\begin{align}
min \sum_{n=1}^{2^{d}}\sum_{p=1}^{\mathcal{P}}\sum_{a=1}^{\mathcal{A}} r_{p,a} \times w_{p,n,a} + \sum_{n=1}^{2^{d}} \beta \times m_{n}
& & & \label{eq:1}
\end{align}
\text{subject to} 
\begin{align}
\sum_{f \in \mathcal{F}}\sum_{c \in {\mathcal{C}_f}}x_{l,n,f,c} = 1 \hspace*{4.2cm} \forall \,\, l \in \{0,\ldots,(d-1)\},  \, n \in \{1,\ldots,2^{l}\}
\label{eq:2}\\
\sum_{n=1}^{2^{d}}\sum_{a=1}^{\mathcal{A}}w_{p,n,a} = 1 \hspace*{9.0cm} \forall \,\, p \in \mathcal{P}
\label{eq:4}\\
 \sum_{a \in \mathcal{A}}z_{n,a} = 1 \hspace*{8.4cm} \forall \,\, n \in \{1,\ldots,2^{d}\}
\label{eq:5}\\
w_{p,n,a} \leq z_{n,a} \hspace*{6.0cm} \forall \,\, p \in \mathcal{P}, \, n \in \{1,\ldots,2^{d}\}, \, a \in \mathcal{A}
\label{eq:6}\\
w_{p,n,a} \leq y_{d,n,p} \hspace*{5.8cm} \forall \,\, p \in \mathcal{P}, \, n \in \{1,\ldots,2^{d}\}, \, a \in \mathcal{A}
\label{eq:7}\\
u_{n} \geq y_{d,n,p} \hspace*{7.5cm} \forall \,\, n \in \{1,\ldots,2^{d}\}, \, p \in \mathcal{P}
\label{eq:8}\\
\sum_{p \in \mathcal{P}}y_{d,n,p} + m_{n} \geq \tau \times u_{n} \hspace*{6.4cm} \forall \,\, n \in \{1,\ldots,2^{d}\}
\label{eq:10}\\
y_{l,n,p} \leq y_{(l-1),max(g_{l,n}, h_{l,n}), p} \hspace*{2.7cm} \forall \,\, l \in \{2,\ldots,d\}, \, n \in \{1,\ldots,2^{l}\}, \, p \in \mathcal{P}
\label{eq:11}\\ 
y_{l,n,p} \leq 1-x_{(l-1), g_{l,n}, f, c} \hspace*{4.2cm} \forall \,\, l  \in \{1,\ldots,d\}, \, n \in \{1,\ldots,2^{l}\},\label{eq:12}\\  p \in \mathcal{P}, \, f \in \mathcal{F}, \, c \in \mathcal{C}_f : g_{l,n} \neq -1 \land v_{p,f} \leq c
\nonumber \\
y_{l,n,p}\leq 1-x_{(l-1), h_{l,n}, f, c}\hspace*{4.2cm}\forall \,\, l \in \{1,\ldots,d\}, n \in \{1,\ldots,2^{l}\},\label{eq:13}\\ p \in \mathcal{P}, \, f \in \mathcal{F}, \, c \in \mathcal{C}_f : h_{l,n} \neq -1 \land v_{p,f} > c
\nonumber \\
x_{l,n,f,c} \in \{0,1\} \hspace*{2.6cm} \forall \,\, l \in \{0,\ldots,(d-1)\}, \, n \in \{1,\ldots,2^{l}\}, \, f \in \mathcal{F}, \, c \in \mathcal{C}
\label{eq:14}\\
y_{l,n,p} \in \{0,1\} \hspace*{4.6cm} \forall \,\, l \in \{1,\ldots,d\}, \, n \in \{1,\ldots,2^{l}\}, \, p \in \mathcal{P}
\label{eq:16}\\
z_{n,a} \in \{0,1\} \hspace*{7.0cm} \forall \,\, n \in \{1,\ldots,2^{d}\}, \, a \in \mathcal{A}
\label{eq:17}\\
u_{n} \in \{0,1\} \hspace*{8.4cm} \forall \,\, n \in \{1,\ldots,2^{d}\}
\label{eq:18}\\
w_{p,n,a} \in \{0,1\} \hspace*{5.6cm} \forall \,\, p \in \mathcal{P}, \, n \in \{1,\ldots,2^{d}\}, \, a \in \mathcal{A}
\label{eq:19}\\
m_{n} \in \mathbb{Z}^{+} \hspace*{8.6cm} \forall n \in \{1,\ldots,2^{d}\}
\label{eq:20}
\end{align}

Equations \ref{eq:2} ensure that each internal node of the tree must have exactly one feature and branching value selected. Each problem instance must be allocated to exactly one leaf node and one algorithm (Equations \ref{eq:4}) and each leaf node must have exactly one associated algorithm (Equations \ref{eq:5}). Inequalities \ref{eq:6} guarantee that the recommended algorithm for a leaf node is the same as the algorithm of the problem instances allocated to this node. Inequalities \ref{eq:7} guarantee that allocations of algorithms to problem instances are performed respecting the leaf node selection for each problem instance.

Constraint set \ref{eq:8} ensures that variables $u_{n}$ are 1 if and only if there is at least one problem instance associated with  leaf node $n$. Constraints \ref{eq:10} ensure that variable $m_{n}$ is set to the number of problem instances missing from the leaf node $n$ to reach the threshold $\tau$. If $m_n = 0$, then the leaf node $n$ contains at least $\tau$ problem instances.

Constraints \ref{eq:11} ensure that any problem instance allocated in a particular node must belong to the associated parent node. Finally, constraints \ref{eq:12} and \ref{eq:13} ensure that problem instances allocated in a particular node respect the feature and branching values selected at the parent node. Constraints \ref{eq:12} are generated when $g_{l,n} \neq -1$ and $v_{p,f} \leq c$ and ensure that problem instance $p$ cannot be allocated at node $n$ of level $l$ ($y_{l,n,p} = 0$), when feature $f$ and branching value $c$ are chosen for its parent node ($x_{(l-1), g_{l,n}, f, c} = 1$). Similarly, Constraints \ref{eq:13} are generated when $h_{l,n} \neq -1$ and $v_{p,f} > c$ and ensure that problem instance $p$ cannot be allocated at node $n$ of level $l$ ($y_{l,n,p} = 0$), when feature $f$ and branching value $c$ are chosen for its parent node ($x_{(l-1), h_{l,n}, f, c} = 1$). Constraints \ref{eq:14}-\ref{eq:20} are related to the domain of the decision variables defined in the model.

\section{VND to accelerate the discovery of better solutions}\label{sec:vnd}

The model proposed in Section \ref{sec:Model} can be optimized by standalone Mixed-Integer Programming (MIP) solvers and in \emph{finite time}  the optimal solution for the ASP will be produced. Despite the continuous evolution of MIP solvers \citep{Johnson2000,Gamrath2013}, the optimization of large MIP models in restricted times, in the general case, is still challenging. Thus, we performed some scalability tests (Section \ref{firstExperiment}) to check how practical it is the use of the our complete model to create optimal decision trees for the ASP in datasets of different sizes in limited times. Since our objective is to produce a method that can tackle large datasets of experiment results, we also propose a mathematical programming heuristic \citep{fischetti2016matheuristics} based on Variable Neighborhood Descent (VND) \citep{vndref1} to speed the production of feasible solutions. VND  is a local search method that consists of exploring the solution space through systematic change of neighborhood structures. Its success is based on the fact that different neighborhood structures do not usually have the same local minimum.

Algorithm \ref{vndalgorithm} shows the pseudo-code for our approach called VND-ASP. The overall method operates as follows: a Greedy Randomized algorithm is employed to generate initial feasible solutions (GRC-ASP). Multiple runs of this constructive algorithm are used to construct an Elite Set of solutions. The best solution of this set is used in our VND local search (MIPSearch), where parts of this solution are fixed and the remaining parts are optimized with the MIP model presented previously. Also, a subset $\mathcal{Q}$ including at most $q$ algorithms is built. It includes all algorithms that appear in the elite set $\mathcal{E}$ and additional algorithms selected with a MIP model as follows: a covering like model to select $q$ algorithms is solved where each instance should be covered by at least $q' < q$ algorithms, minimizing the cost of covering each problem instance with the selected algorithm.

\begin{algorithm}[H]
\caption{VND-ASP ($r$, $\mathcal{h}$, $\mathcal{l}$, $D$, $d$, $\alpha$, $\mathcal{m}$, $\mathcal{n}$, $\mathcal{Q}$, $\mathcal{A}$, $\mathcal{P}$, $\mathcal{q}$, $\mathcal{q'}$)}\label{vndalgorithm}
\textbf{Input}. matrix $r$: algorithm performance matrix; set $D$: all different branching values for all features ($\mathcal{F}\times\mathcal{C}_f$); set $\mathcal{A}$: set of algorithms; set $\mathcal{P}$: set of problems.

\textbf{Parameters}. $\mathcal{h}$: matheuristic execution timeout; $\mathcal{l}$: MIP search execution timeout; $d$: maximum depth; $\alpha$: represents a continuous value between 0.1 and 1.0 that controls the greedy or random construction of the solution; $\mathcal{m}$: maximum number of trees; $\mathcal{n}$: maximum number of iterations without improvement; $\mathcal{q}$: defines the number of algorithms of the set $\mathcal{A}$, $\mathcal{q'}$: minimum number of algorithms to cover each problem instance.
\begin{algorithmic}[1]
\STATE{$\mathcal{E}$ $\gets \{\}$; $\mathcal{T}$ $ \gets \{\}$; $st \gets $ time()} 
\STATE{$\mathcal{Q}$ $\gets$ algsubset ($r$, $\mathcal{A}$, $\mathcal{P}$, $\mathcal{E}$, $\mathcal{q}$, $\mathcal{q'}$)}
\STATE{GRC-ASP ($\mathcal{P}$, $\mathcal{A}$, $r$, $\mathcal{T}$, 0, $D$, $d$, 1.0, $\mathcal{m}$, $\mathcal{E}$, $\mathcal{Q}$)}
\STATE{$\mathcal{i}$ $\gets$ 0}
\WHILE{($\mathcal{i}$ $<$ $\mathcal{n}$)}
    \STATE{$\mathcal{T}$ $ \gets \{\}$}
    \STATE{GRC-ASP ($\mathcal{P}$, $\mathcal{A}$, $r$, $\mathcal{T}$, $\mathcal{i}$, $D$, $d$, rand (0.1, 1.0), $\mathcal{m}$, $\mathcal{E}$, $\mathcal{Q}$)}
    \IF{(PerformanceDegradation ($\mathcal{T}$) $<$ HigherPerformanceDegradation ($\mathcal{E}$))}
	        \IF{(PerformanceDegradation ($\mathcal{T}$) $<$ LowerPerformanceDegradation ($\mathcal{E}$))}
	            \STATE{$\mathcal{i}$ $\gets$ -1}
	        \ENDIF
	        \STATE{$\mathcal{i}$ $\gets$ $\mathcal{i}$ + 1}
	        \IF{($\mathcal{T}$ $\notin$ $\mathcal{E}$)}
	            \IF{($|\mathcal{E}|$ $<$ $\mathcal{m}$)}
	                \STATE{$\mathcal{E}$ $\gets$ $\mathcal{E}$ $\cup$ $\{\mathcal{T}\}$}
	            \ELSE
	                \STATE{$\mathcal{s}$ $\gets$ SimilarityTrees ($\mathcal{E}$, $\mathcal{T}$); $\mathcal{E}_{\mathcal{s}}$ $\gets$ $\mathcal{T}$}
	            \ENDIF
	           \STATE{$\mathcal{Q}$ $\gets$ $\mathcal{Q}$ $\cup$ AlgorithmsLeafNodes($\mathcal{T}$)}
	        \ENDIF
	   \ELSE
	        \STATE{$\mathcal{i}$ $\gets$ $\mathcal{i}$ + 1}
        \ENDIF
\ENDWHILE
\STATE{Let $s$ be the best solution of the set $\mathcal{E}$}
\STATE{Let $\mathcal{J}$ be the list of all subproblems in all neighborhoods}
\STATE{$\mathcal{J}$ $\gets$ Shuffle(); $k \gets 1$; $ft \gets $ time(); $et \gets ft - st$}
\WHILE{($k \leq |\mathcal{J}|$ $\AND$ $\mathcal{et}$ $\leq$ $\mathcal{h}$)}
    \STATE{$s^{'} \gets$ MIPSearch ($s^{}$, $k$, $\mathcal{l}$, $\mathcal{Q}$. $\mathcal{J}$)}
    \IF{$(f(s^{'}) < f(s))$}
    	\STATE{$s \gets s^{'}$; $k \gets 1$; $\mathcal{J}$ $\gets$ Shuffle($\mathcal{J}$)}
        \ELSE
        	\STATE{$k \gets k + 1$}
    \ENDIF
    \STATE{$ft \gets $ time(); $et \gets ft - st$}
\ENDWHILE
\STATE{Return $s$;}
\end{algorithmic}
\end{algorithm}

Our matheuristic has the following parameters: $\mathcal{h}$ indicates the time limit for running the entire algorithm; $\mathcal{l}$ indicates the time limit for a single exploration of a sub-problem in the MIP based neighborhood search; $d$ is the maximum tree depth; parameter $\alpha \in [0, 1]$ controls the randomization of the greedy algorithm; parameter $\mathcal{m}$ controls the maximum size of the set of trees $\mathcal{E}$; executions of the GRC-ASP algorithm are restricted to at most $n$ iterations without updates in the elite set. Set $D$ represents the different features and cut-off points of the problem instances.

The list $\mathcal{J}$ contains all sub-problems that can be obtained by fixing solution components in all neighborhoods. We shuffle these sub-problems in a list so that there is no priority for searching first in one neighborhood relative to another. This strategy is inspired by \citep{SOUZA20101041}, where several neighborhood orders were in tested in a VND algorithm and the randomized order obtained better results.  Whenever the incumbent solution is updated, the list $\mathcal{J}$ is shuffled again and the algorithm starts to explore it from the beginning.

In the  following sections we describe in more details the algorithm used to generate initial feasible solutions (Section \ref{inputvnd}) and the MIP based neighborhoods employed in our algorithm (Section \ref{neighborhoods}).

\subsection{Constructive Algorithm}\label{inputvnd}

The initial solution $s$ is obtained from the best solution of the set of trees $\mathcal{E}$. This set is obtained using a hybrid approach inspired by \cite{Quinlan:1993:CPM:152181}'s C4.5 algorithm to generate a decision tree and the Greedy Randomized Constructive (GRC) \citep{graspref} search. GRC searches to a certain randomness in the greedy criterion adopted by the C4.5 algorithm. Algorithm \ref{c45grasp} shows the hybrid approach. Lines 7-16 of Algorithm \ref{c45grasp} (GRC-ASP) show the adaptation made in the C4.5 algorithm for use of the restricted candidate list of the GRC search.

Another adaptation considers the metric to split the nodes. Algorithm C4.5 uses \textbf{information gain} metric. This metric aims to choose the attribute that minimizes the impurity of the data. In a data set, it is a measure of the lack of homogeneity of the input data in relation to its classification. In our case, we used the \textbf{performance degradation} metric to split the nodes. This metric searches for the attribute that minimizes the degradation of performance obtained using the recommended algorithm from the ideal performance for each problem $p$.

\begin{algorithm}[H]
\caption{GRC-ASP ($\mathcal{P}$, $\mathcal{A}$, $r$, $\mathcal{T}$, $\mathcal{i}$, $D$, $d$, $\alpha$, $\mathcal{m}$, $\mathcal{E}$, $\mathcal{Q}$)}\label{c45grasp}
\begin{algorithmic}[1]
    \IF{(current depth $=$ $d$)}
    \STATE{terminate}
    \ENDIF
    \FORALL{$(f,c)$ in $D$}
    	\STATE{$\kappa_{f,c}$ $\gets$ PerformanceDegradationTree ($\mathcal{T}$, $r$, $f$, $c$)}
    \ENDFOR
    \STATE{RCL $\gets \emptyset$}
    \STATE{$\underline{a}$ $\gets$ max($\kappa_{f,c}$)}
    \STATE{$\overline{a}$ $\gets$ min($\kappa_{f,c}$)}
    \STATE{$pdt$ $\gets$ $\underline{a}$ + $\alpha$ $\times$ ($\overline{a}$ - $\underline{a}$)}
    \FORALL{$(f,c)$ in $D$}
    	\IF{$\kappa_{f,c}$ $\leq$ $pdt$}
        	\STATE{RCL $\gets$ RCL $\cup$ $(f,c)$}
        \ENDIF
    \ENDFOR
    \STATE{$a_{rcl}$ = Randomly select a $(f,c)$ from the RCL list.}
    \STATE{$\mathcal{t}$ = Create a decision node in $\mathcal{T}$ that tests $a_{rcl}$ in the root}
    \STATE{$D_{l}$ = Induced sub-dataset of $r$ whose value of feature $f$ are less than or equal to the cut-off point $a_{rcl}$}
    \STATE{$D_{r}$ = Induced sub-dataset of $r$ whose value of feature $f$ are greater cut-off point $a_{rcl}$}
	\STATE{$\mathcal{T}_{l}$ = GRC-ASP ($\mathcal{P}$, $\mathcal{A}$, $r$, $\mathcal{T}$, $\mathcal{i}$, $D_{l}$, $d$, $\alpha$, $\mathcal{m}$, $\mathcal{n}$, $\mathcal{E}$, $\mathcal{Q}$)}
	\STATE{Attach $\mathcal{T}_{l}$ to the corresponding branch of $\mathcal{t}$}
	\STATE{$\mathcal{T}_{r}$ = GRC-ASP ($\mathcal{P}$, $\mathcal{A}$, $r$, $\mathcal{T}$, $\mathcal{i}$, $D_{r}$, $d$, $\alpha$, $\mathcal{m}$, $\mathcal{n}$, $\mathcal{E}$, $\mathcal{Q}$)}
	\STATE{Attach $\mathcal{T}_{r}$ to the corresponding branch of $\mathcal{t}$}
\end{algorithmic}
\end{algorithm}

\subsection{Neighborhoods}\label{neighborhoods}
\begin{figure}[H]
  \centerline{\includegraphics[width=16.5cm]{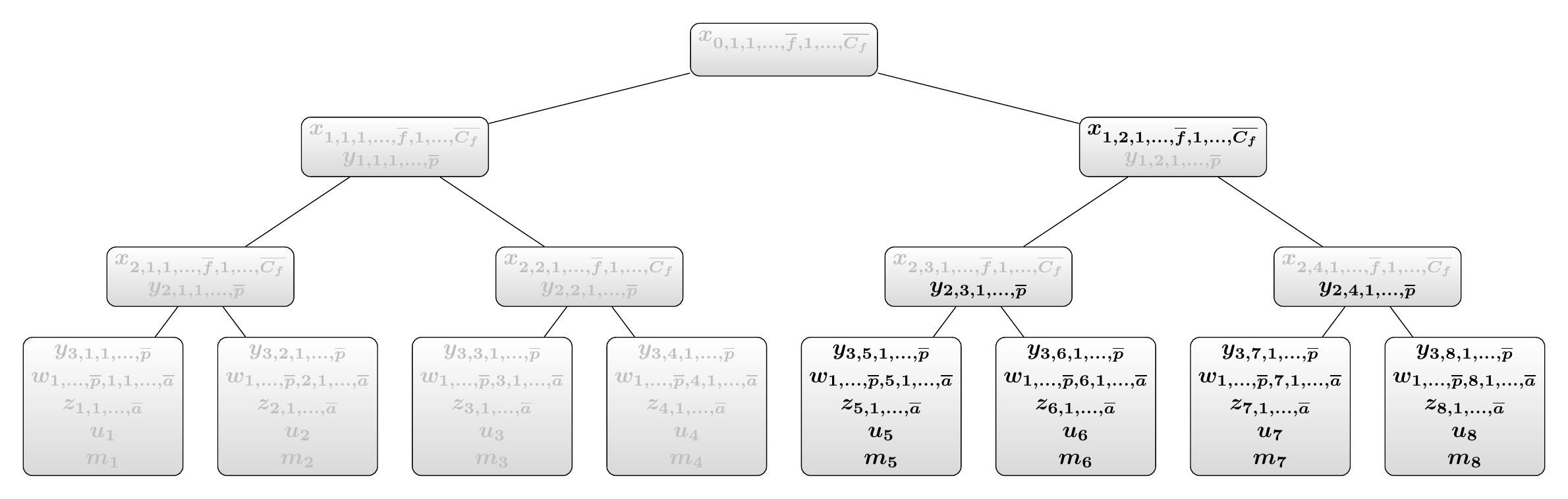}}
\caption{Example of neighborhood $\mathcal{N}_1$ variables: variables highlighted in gray are fixed and variables highlighted in black will be optimized.}
\label{neighsfigure2}
\vspace{0.3cm}
\centerline{\includegraphics[width=16.5cm]{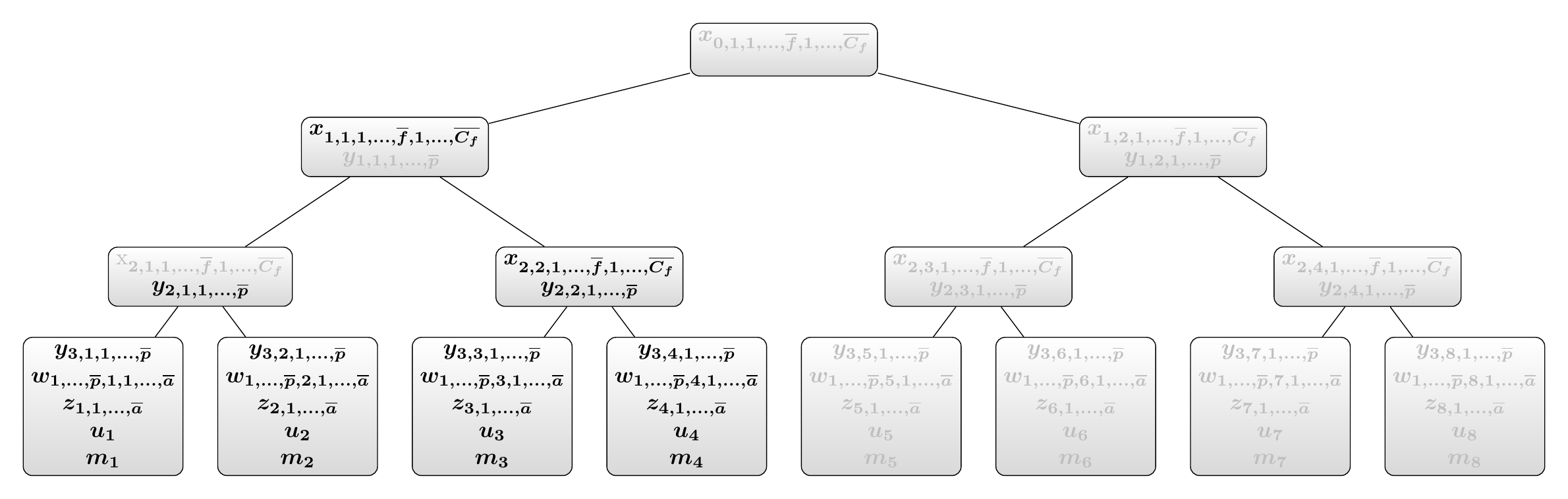}}
\caption{Example of neighborhood $\mathcal{N}_2$ variables: variables highlighted in gray are fixed and variables highlighted in black will be optimized.}
\label{neighsfigure3}
\vspace{0.3cm}
  \centerline{\includegraphics[width=16.5cm]{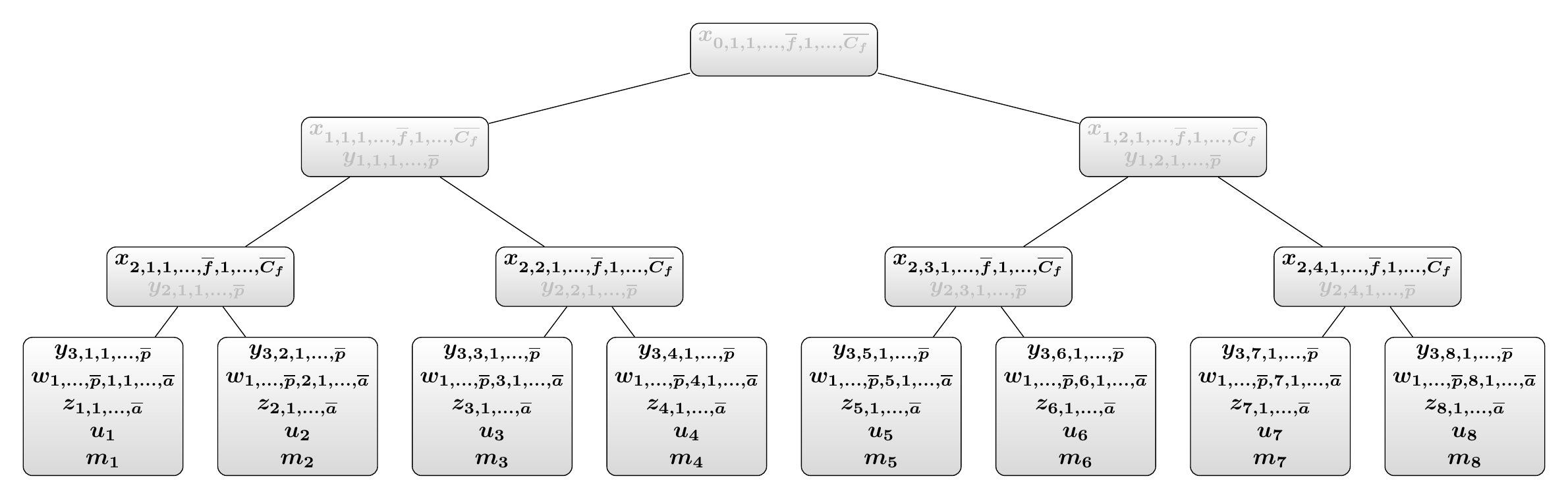}}
\caption{Example of neighborhood $\mathcal{N}_3$ variables: variables highlighted in gray are fixed and variables highlighted in black will be optimized.}
\label{neighsfigure4}
\end{figure}

\begin{figure}[H]
  \centerline{\includegraphics[width=16.5cm]{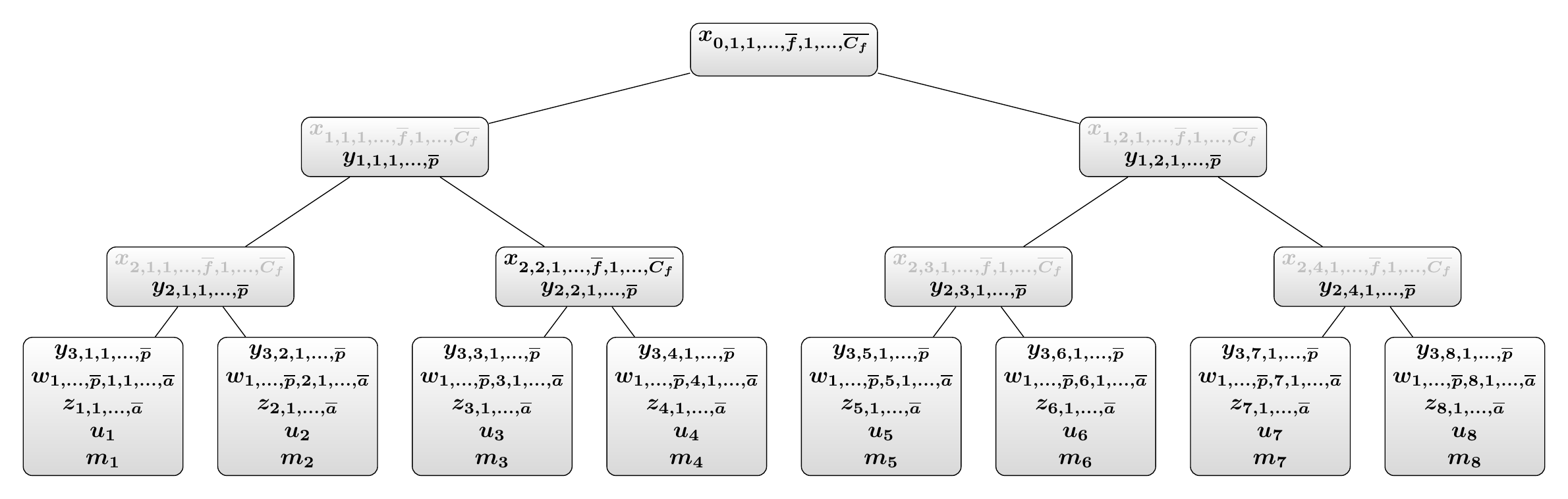}}
\caption{Example of neighborhood $\mathcal{N}_4$ variables: variables highlighted in gray are fixed and variables highlighted in black will be optimized.}
\label{neighsfigure5}
\vspace{0.3cm}
  \centerline{\includegraphics[width=16.5cm]{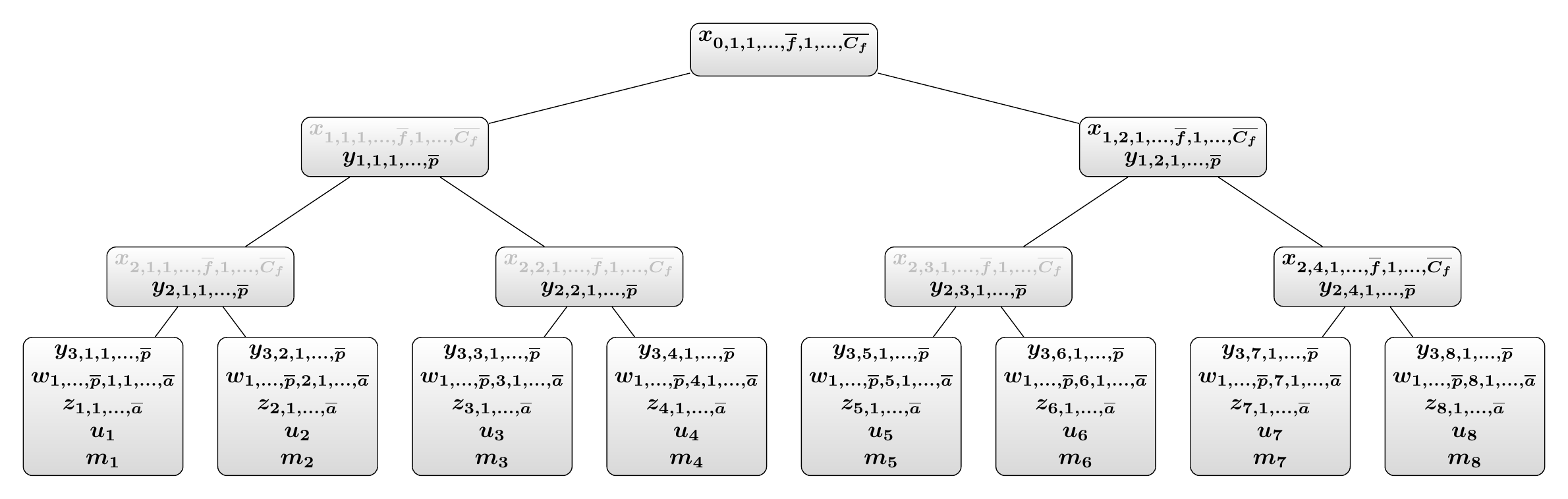}}
\caption{Example of neighborhood $\mathcal{N}_5$ variables: variables highlighted in gray are fixed and variables highlighted in black will be optimized.}
\label{neighsfigure6}
\end{figure}

Since optimizing the complete MIP model may be too expensive, five neighborhoods have been designed to be employed in a fix-and-optimize context. Each neighborhood defines a set of subproblems to be optimized. These neighborhoods are explained in what follows, together with examples shown in Figs. \ref{neighsfigure2}-\ref{neighsfigure6}. Examples consider a decision tree with 4 levels: variables in gray are fixed and variables highlighted in black will be optimized. These neighborhoods are explored in a  Variable Neighborhood Descent using the MIP solver in a fix-and-optimize strategy. Not only the neighborhoods, but the sub-problems of all neighboorhoods, are explored in a random order. 

We will explain how the decision variables $x_{l,n,f,c}$, $y_{l,n,p}$ and $z_{n,a}$ are optimized in each of the neighborhoods. The following neighborhoods were developed: 
\begin{itemize}
\item \textbf{Neighborhood $\mathcal{N}_1$}:
optimizes the selection of the \textbf{feature} and the \textbf{cut-off point} of an internal node and consequently optimizes the allocation of problems in child nodes as well as optimizes the choice of the recommended algorithm in the respective leaf nodes.

In the example of Figure \ref{neighsfigure2}, we consider optimizing the feature and cut-off point of internal node 2 at level 1 of the tree (binary variables $x_{1,2,1,\ldots,\overline{f},1,\ldots,\overline{C_{f}}}$). Since these variables determine the problems that will be allocated to the left and right child nodes, the binary variables $y_{2,3,1,\ldots,\overline{p}}$, $y_{2,4,1,\ldots,\overline{p}}$, $y_{3,5,1,\ldots,\overline{p}}$, $y_{3,6,1,\ldots,\overline{p}}$, $y_{3,7,1,\ldots,\overline{p}}$ and $y_{3,8,1,\ldots,\overline{p}}$ will also be optimized. Moreover, the recommended algorithm to the problems allocated in all child nodes in relation to the chosen node - internal node 2 of level 1 of the tree - which are leaf nodes should also be optimized. In our example, these would be the binary variables $z_{5,1,\ldots,\overline{a}}$, $z_{6,1,\ldots,\overline{a}}$, $z_{7,1,\ldots,\overline{a}}$ and $z_{8,1,\ldots,\overline{a}}$ of the leaf nodes 5,\ldots,8 of level 3 of the tree.

\item \textbf{Neighborhood $\mathcal{N}_2$}: optimizes the selection of the \textbf{feature} and the \textbf{cut-off point} of an internal node (this node cannot be the root node) and optimizes the choice of the \textbf{feature} and the \textbf{cut-off point} of the associated \textbf{parent node}, it consequently optimizes the allocation of problems in child nodes of associated parent node as well as the choice of the recommended algorithm in the respective leaf nodes.

In the example of Figure \ref{neighsfigure3}, we consider optimizing the feature and cut-off point of internal node 2 of level 2 of the tree (binary variables $x_{2,2,1,\ldots,\overline{f},1,\ldots,\overline{C_{f}}}$) and optimizing the feature and cut-off point of associated parent node (binary variables $x_{1,1,1,\ldots,\overline{f},1,\ldots,\overline{C_{f}}}$). Since these variables determine the problems that will be allocated to the left and right child nodes, the binary variables $y_{2,1,1,\ldots,\overline{p}}$, $y_{2,2,1,\ldots,\overline{p}}$, $y_{3,1,1,\ldots,\overline{p}}$, $y_{3,2,1,\ldots,\overline{p}}$, $y_{3,3,1,\ldots,\overline{p}}$ and $y_{3,4,1,\ldots,\overline{p}}$ will also be optimized. Moreover, the recommended algorithm to execute the problems allocated in all child nodes in relation to the associated parent node - internal node 1 of level 1 of the tree - which are leaf nodes should also be optimized. In our example, these would be the binary variables $z_{1,1,\ldots,\overline{a}}$, $z_{2,1,\ldots,\overline{a}}$, $z_{3,1,\ldots,\overline{a}}$ and $z_{4,1,\ldots,\overline{a}}$ of the leaf nodes 1,\ldots,4 at level 3 of the tree.

\item \textbf{Neighborhood $\mathcal{N}_3$}: optimizes the selection of the \textbf{feature} and the \textbf{cut-off point} of all nodes at one level (the level of the root node cannot be chosen) of the decision tree. Consequently optimizes the allocation of the problems in the nodes of the subsequent levels to the chosen level, it in addition to the choice of the recommended algorithm in the respective leaf nodes.

In the example of Figure \ref{neighsfigure4}, we consider optimizing the feature and cut-off point of all nodes of level 2 of the tree (binary variables $x_{2,1,1,\ldots,\overline{f},1,\ldots,\overline{C_{f}}}$, $x_{2,2,1,\ldots,\overline{f},1,\ldots,\overline{C_{f}}}$, $x_{2,3,1,\ldots,\overline{f},1,\ldots,\overline{C_{f}}}$ and $x_{2,4,1,\ldots,\overline{f},1,\ldots,\overline{C_{f}}}$). Since these variables determine the problems that will be allocated at subsequent levels, the binary variables $y_{3,1,1,\ldots,\overline{p}}$, $y_{3,2,1,\ldots,\overline{p}}$, $y_{3,3,1,\ldots,\overline{p}}$, $y_{3,4,1,\ldots,\overline{p}}$, $y_{3,5,1,\ldots,\overline{p}}$, $y_{3,6,1,\ldots,\overline{p}}$, $y_{3,7,1,\ldots,\overline{p}}$ and $y_{3,8,1,\ldots,\overline{p}}$ will also be optimized. In addition, the recommended algorithm to execute the problems allocated in all leaf nodes should also be optimized. In our example, these would be binary variables $z_{1,1,\ldots,\overline{a}}$, $z_{2,1,\ldots,\overline{a}}$, $z_{3,1,\ldots,\overline{a}}$, $z_{4,1,\ldots,\overline{a}}$, $z_{5,1,\ldots,\overline{a}}$, $z_{6,1,\ldots,\overline{a}}$, $z_{7,1,\ldots,\overline{a}}$, and $z_{8,1,\ldots,\overline{a}}$ of leaf nodes 1,\ldots,8 at level 3 of the tree.

\item \textbf{Neighborhood $\mathcal{N}_4$}: optimizes the selection of the \textbf{feature} and the \textbf{cut-off point} of the root node and the choice of the \textbf{feature} and the \textbf{cut-off point} of an internal node, so that this node is at least at the third level of the tree ($l = 2$). Consequently it optimizes the allocation of problems in all nodes of the tree as well as the choice of the recommended algorithm in the respective leaf nodes.

In the example of Figure \ref{neighsfigure5}, we consider optimizing the feature and cut-off point of both the root node (binary variables $x_{0,1,1,\ldots,\overline{f},1,\ldots,\overline{C_{f}}}$) and optimizing internal node 2 at level 2 of the tree (binary variables $x_{2,2,1,\ldots,\overline{f},1,\ldots,\overline{C_{f}}}$). Since these variables determine the problems that will be allocated to all other nodes of the tree, binary variables $y_{1,1,1,\ldots,\overline{p}}$, $y_{1,2,1,\ldots,\overline{p}}$, $y_{2,1,1,\ldots,\overline{p}}$, $y_{2,2,1,\ldots,\overline{p}}$, $y_{2,3,1,\ldots,\overline{p}}$, $y_{2,4,1,\ldots,\overline{p}}$, $y_{3,1,1,\ldots,\overline{p}}$, $y_{3,2,1,\ldots,\overline{p}}$, $y_{3,3,1,\ldots,\overline{p}}$, $y_{3,4,1,\ldots,\overline{p}}$, $y_{3,5,1,\ldots,\overline{p}}$, $y_{3,6,1,\ldots,\overline{p}}$, $y_{3,7,1,\ldots,\overline{p}}$ and $y_{3,8,1,\ldots,\overline{p}}$ will also be optimized. In addition, the recommended algorithm to execute the problems allocated to all leaf nodes should also be optimized. In our example, these would be binary variables $z_{1,1,\ldots,\overline{a}}$, $z_{2,1,\ldots,\overline{a}}$, $z_{3,1,\ldots,\overline{a}}$, $z_{4,1,\ldots,\overline{a}}$, $z_{5,1,\ldots,\overline{a}}$, $z_{6,1,\ldots,\overline{a}}$, $z_{7,1,\ldots,\overline{a}}$, and $z_{8,1,\ldots,\overline{a}}$ of leaf nodes 1,\ldots,8 at level 3 of the tree.

\item \textbf{Neighborhood $\mathcal{N}_5$}: optimizes the selection of the \textbf{feature} and the \textbf{cut-off point} of a particular path from the root node to one of the tree's leaf nodes.
Consequently it both optimizes the allocation of problems in all nodes of the tree and the choice of the recommended algorithm in the respective leaf nodes.

In the example of Figure \ref{neighsfigure6}, we consider the path from the root node to leaf node 8. We consider optimizing the feature and cut-off point of both the root node (binary variables $x_{0,1,1,\ldots,\overline{f},1,\ldots,\overline{C_{f}}}$), internal node 2 at level 1 of the tree (binary variables $x_{1,2,1,\ldots,\overline{f},1,\ldots,\overline{C_{f}}}$), and internal node 4 at level 2 of the tree (binary variables $x_{2,4,1,\ldots,\overline{f},1,\ldots,\overline{C_{f}}}$). Since these variables determine the problems that will be allocated to all other nodes of the tree, binary variables $y_{1,1,1,\ldots,\overline{p}}$, $y_{1,2,1,\ldots,\overline{p}}$, $y_{2,1,1,\ldots,\overline{p}}$, $y_{2,2,1,\ldots,\overline{p}}$, $y_{2,3,1,\ldots,\overline{p}}$, $y_{2,4,1,\ldots,\overline{p}}$, $y_{3,1,1,\ldots,\overline{p}}$, $y_{3,2,1,\ldots,\overline{p}}$, $y_{3,3,1,\ldots,\overline{p}}$, $y_{3,4,1,\ldots,\overline{p}}$, $y_{3,5,1,\ldots,\overline{p}}$, $y_{3,6,1,\ldots,\overline{p}}$, $y_{3,7,1,\ldots,\overline{p}}$ and $y_{3,8,1,\ldots,\overline{p}}$ will also be optimized. In addition, the recommended algorithm to execute the problems allocated to all leaf nodes should also be optimized. In our example, these would be binary variables $z_{1,1,\ldots,\overline{a}}$, $z_{2,1,\ldots,\overline{a}}$, $z_{3,1,\ldots,\overline{a}}$, $z_{4,1,\ldots,\overline{a}}$, $z_{5,1,\ldots,\overline{a}}$, $z_{6,1,\ldots,\overline{a}}$, $z_{7,1,\ldots,\overline{a}}$, and $z_{8,1,\ldots,\overline{a}}$ of leaf nodes 1,\ldots,8 at level 3 of the tree.

\end{itemize}

\section{Experiments}\label{sec:Experiments} 
The computational experiments are divided into two groups: the first group (Section \ref{firstExperiment}) is dedicated to check the scalability of our integer programming model using a standalone MIP solver in datasets with different numbers of problem instances and algorithms. The second group of experiments (Section \ref{secondExperiment}) uses cross-validation and divides the complete base into 10 partitions, where training and test partitions are used. Section \ref{initDef} presents the initial configurations for these two experiments. 

\subsection{Computational experiments dataset}\label{initDef}
This section describes how the experiments database was created. Details of the sets of problem instances and algorithms will be presented in Sections \ref{problemInstances} and \ref{availableAlgorithms}.
\subsubsection{Problem instances}\label{problemInstances}
Computational experiments were performed for a diverse set of 1004 problem instances including the MIPLIB 3, 2003, 2010 and 2017 \citep{koch2011miplib} benchmark sets. Additional instances from Nurse Rostering \citep{Santos2016}, School Timetabling \citep{Fonseca2017} and Graph Drawing \citep{esilva2017} were also included. We extracted 37 features ($|\mathcal{F}| = 37$) associated to variables, constraints and coefficients in the constraint matrix for problem instances. These features are similar to the ones used in \citep{hutter201479} with the notable exception that features that are  computationally expensive to extract were discarded to ensure that our approach would incur no overhead when incorporated into an algorithm. When building the problem instances dataset, special care was taken to ensure that no application was over-represented. Table \ref{featuresInstances} shows the minimum (min), maximum (max), average (avg) and standard deviation (sd) of each feature over the complete set of problem instances. The density feature was computed as $(\frac{nz}{rows\times cols})\times 100$.

\begin{table}[H]
\caption{Distribution of problem instances according to features\label{featuresInstances}}
\begin{tabular*}{\hsize}{@{}@{\extracolsep{\fill}}|c|rrrr|c|rrrr|@{}}
\hline
\scriptsize{\textbf{feature}} & \scriptsize{\textbf{min}} & \scriptsize{\textbf{max}} & \scriptsize{\textbf{avg}} &\scriptsize{ \textbf{sd}} & \scriptsize{\textbf{feature}} & \scriptsize{\textbf{min}} & \scriptsize{\textbf{max}} & \scriptsize{\textbf{avg}} & \scriptsize{\textbf{sd}}\\
\hline 
\scriptsize{cols} & \scriptsize{3}& \scriptsize{2277736}& \scriptsize{35368.04}& \scriptsize{120402.72}&
\scriptsize{rflowint}  & \scriptsize{0}&\scriptsize{120201} & \scriptsize{428.23}& \scriptsize{4052.12}\\
\scriptsize{bin} & \scriptsize{0}& \scriptsize{2277736}& \scriptsize{25402.59}& \scriptsize{103781.75}&
\scriptsize{rflowmx}  & \scriptsize{0}& \scriptsize{410733}& \scriptsize{1813.33}&\scriptsize{17306.95} \\
\scriptsize{int} & \scriptsize{0}& \scriptsize{440899}& \scriptsize{2301.02}& \scriptsize{16289.21}& \scriptsize{rvbound}  &\scriptsize{0} & \scriptsize{0}& \scriptsize{0}& \scriptsize{0}\\
\scriptsize{cont} & \scriptsize{0}& \scriptsize{799416}& \scriptsize{7664.43}& \scriptsize{47456.96}&
\scriptsize{rother}  &\scriptsize{0} &\scriptsize{2365080} & \scriptsize{26491.79}& \scriptsize{110176.92}\\
\scriptsize{objMin} & \scriptsize{-1.72E+11}& \scriptsize{5084550}& \scriptsize{-1.72E+08}& \scriptsize{5.431E+09}&
\scriptsize{rhsMin}  &\scriptsize{-1E+100} & \scriptsize{367127}& \scriptsize{-2.98E+97}& \scriptsize{5.45E+98}\\
\scriptsize{objMax} & \scriptsize{-400071}& \scriptsize{1.13E+09}& \scriptsize{4398763.75}& \scriptsize{5.9E+07}&
\scriptsize{rhsMax}  &\scriptsize{-6.47} & \scriptsize{1E+100}& \scriptsize{1.39E+98}&\scriptsize{1.17E+99} \\
\scriptsize{objAv} & \scriptsize{-2.54E+10}& \scriptsize{5.00E+07}& \scriptsize{-2.52E+07}& \scriptsize{8E+08}&
\scriptsize{rhsAv}  & \scriptsize{-5.24E+95}& \scriptsize{5.12E+098}& \scriptsize{5.13E+95}& \scriptsize{1.61E+97}\\
\scriptsize{objMed} & \scriptsize{-1.55E+08}& \scriptsize{40212300}& \scriptsize{-192578.19}& \scriptsize{7E+06}&
\scriptsize{rhsMed}  & \scriptsize{-29961800}&\scriptsize{999949} & \scriptsize{-33807.20}& \scriptsize{979700.89}\\
\scriptsize{objAllInt} & \scriptsize{0}& \scriptsize{1}& \scriptsize{0.70}& \scriptsize{0.46}& 
\scriptsize{rhsAllInt} & \scriptsize{0}&\scriptsize{1} & \scriptsize{0.82}&\scriptsize{0.39} \\
\scriptsize{objRatioLSA} & \scriptsize{-1}& \scriptsize{2.24E+12}& \scriptsize{2.448E+09}&
\scriptsize{7.081E+10}&  \scriptsize{rhsRatioLSA} &\scriptsize{-1} &\scriptsize{1.01E+103} & \scriptsize{2.02E+100}&\scriptsize{4.50E+101} \\
\scriptsize{rows} & \scriptsize{1}& \scriptsize{2897380}& \scriptsize{38440.14}& \scriptsize{146476.08}&
\scriptsize{equalities} & \scriptsize{0}& \scriptsize{416449}& \scriptsize{5745.37}& \scriptsize{27584.05} \\
\scriptsize{rpart} &\scriptsize{0} &\scriptsize{18431} &
\scriptsize{255.05}& \scriptsize{1084.42}& \scriptsize{nz} & \scriptsize{3}& \scriptsize{27329856}& \scriptsize{390757.39}& \scriptsize{1567477.92} \\
\scriptsize{rpack} & \scriptsize{0}& \scriptsize{773664}&\scriptsize{4598.49} &\scriptsize{44125.82} &
\scriptsize{aMin}  & \scriptsize{-4E+09}&\scriptsize{1} & \scriptsize{-4523186.29}& \scriptsize{1E+08}\\
\scriptsize{rcov} & \scriptsize{0}&\scriptsize{88452} &
\scriptsize{381.60}& \scriptsize{3481.73}& \scriptsize{aMax}  &\scriptsize{-1} & \scriptsize{370795000}& \scriptsize{1891549.90}& \scriptsize{2E+07}\\
\scriptsize{rcard} &\scriptsize{0} & \scriptsize{430}&
\scriptsize{9.95}& \scriptsize{33.88}& \scriptsize{aAv}  &\scriptsize{-107447000}&\scriptsize{1148410} & \scriptsize{-101191.22}& \scriptsize{3391461.70}\\
\scriptsize{rknp} & \scriptsize{0}&\scriptsize{103041} &
\scriptsize{189.01}& \scriptsize{3356.63}& \scriptsize{aMed}  & \scriptsize{-41014}& \scriptsize{10000}& \scriptsize{-25.70}& \scriptsize{1353.80}\\
\scriptsize{riknp} &\scriptsize{0} & \scriptsize{547200}& \scriptsize{2062.42}& \scriptsize{29593.52}&
\scriptsize{aAllInt}  & \scriptsize{0}&\scriptsize{1} & \scriptsize{0.69}& \scriptsize{0.46}\\
\scriptsize{rflowbin} & \scriptsize{0}& \scriptsize{381806}& \scriptsize{2210.27}&\scriptsize{19281.02} &
\scriptsize{aRatioLSA}  & \scriptsize{1}&\scriptsize{5.787E+12} & \scriptsize{6.117E+09}&\scriptsize{1.824E+11} \\
\scriptsize{density} & \scriptsize{0.0001819}& \scriptsize{100}& \scriptsize{5.44}&\scriptsize{17.50} &   & & & & \\
\hline
\end{tabular*}
\end{table}

\begin{figure}[H]
  \centerline{\includegraphics[width=13cm]{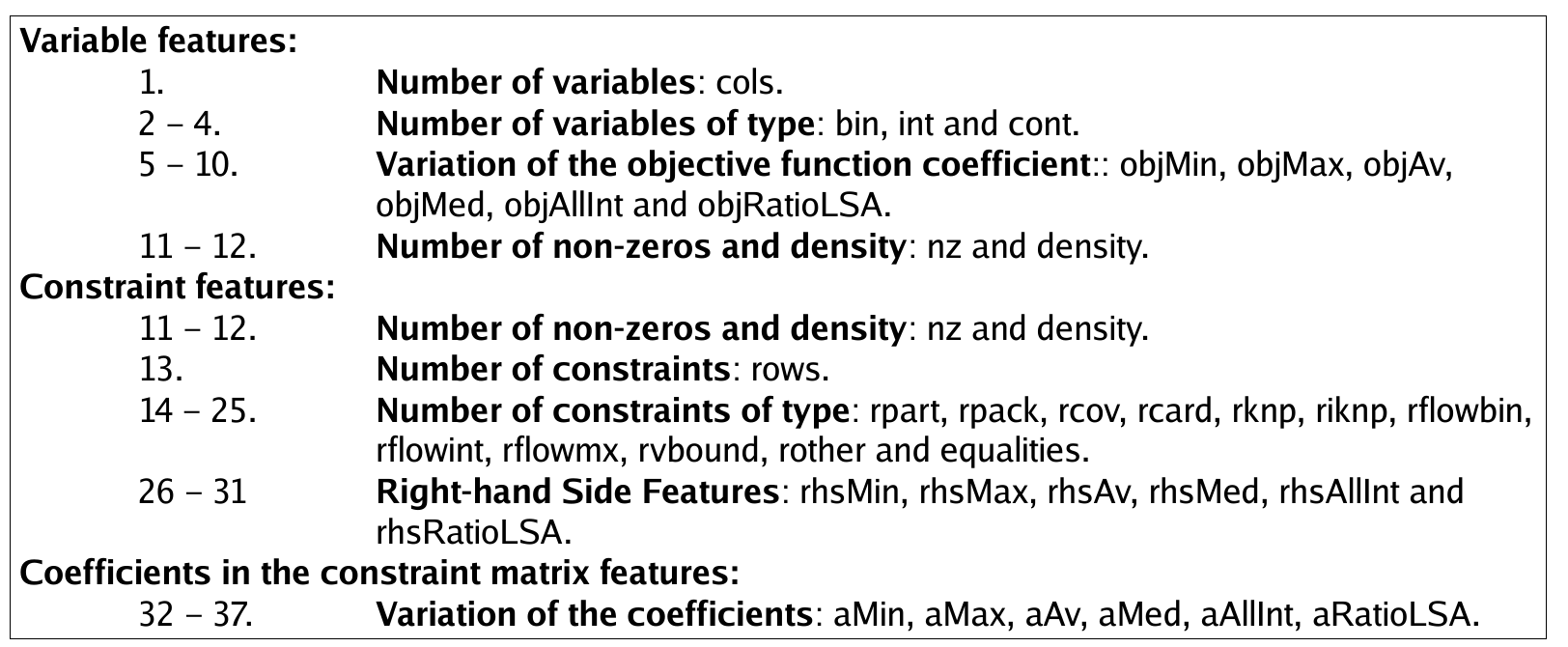}}
\caption{Features of problem instances of Algorithm Selection Problem: variables, constraints and coefficients in the constraint matrix.}
\label{featureLegal}
\end{figure}

Fig. \ref{featureLegal} summarizes the 37 features for ASP, grouped by features related to variables, constraints and coefficients in the constraint matrix.

\subsubsection{Available algorithms}\label{availableAlgorithms}

The definition of the solution method for the LP solver in CBC involves selecting the algorithm, such as dual simplex or the barrier and defining several parameters, such as the perturbation value and the pre-solve effort.
Overall 532 different algorithm configurations were evaluated for each one of the 1004 problem instances . A timeout $T$ = 4000 was set for each execution. The computational results matrix $r$ was filled with the execution time for regular executions, i.e. executions that finished before the time limit and provided correct results. Executions for a given problem instance $p$ and algorithm $a$ that crashed, exceeded the time limit or produced wrong results were penalized by setting $r_{pa} = 8000$. This large batch of experiments was executed in computers with 32 Gb of RAM and 10 Intel \textregistered i9-7900X processing cores. Tasks where scheduled in parallel (7 threads simultaneously)  with the GNU Parallel package \citep{tange2011gnu}. 
Table \ref{parametersAlgorithms} shows the algorithms and parameter values evaluated. This experiment to generate the experimental results dataset produced some interesting results itself: a new better single parameter setting was discovered that decrease the solution time by 22\% in average, a remarkable improvement considering that CLP is already the fastest open source linear programming solver. 
\begin{table}[ht]
\caption{Algorithms and some parameters values evaluated\label{parametersAlgorithms}}
\begin{tabular*}{\hsize}{@{}@{\extracolsep{\fill}}|p{0.6cm}|p{0.9cm}|p{1.1cm}|p{0.6cm}|p{0.8cm}|p{0.7cm}|p{0.8cm}|p{1.0cm}|p{0.4cm}|p{0.7cm}|p{0.7cm}|p{0.7cm}|p{0.7cm}|p{0.5cm}|@{}}
\hline
\scriptsize{\textbf{$\mathcal{A}$}} & \scriptsize{\textbf{idiot}} & \scriptsize{\textbf{crash}} & \scriptsize{\textbf{dualize}} & \scriptsize{\textbf{pertv}} & \scriptsize{\textbf{sprint}} & \scriptsize{\textbf{primalp}} & \scriptsize{\textbf{psi}}& \scriptsize{\textbf{scal}} & \scriptsize{\textbf{passp}}
& \scriptsize{\textbf{subs}} & \scriptsize{\textbf{perturb}}& \scriptsize{\textbf{presolve}} & \scriptsize{\textbf{spars}}\\
\hline
\scriptsize{primals} & \scriptsize{\{3,4,5,6} & \scriptsize{\{idiot1$\ldots$7,} & \scriptsize{\{0,1,2,} & \scriptsize{\{-3500,} & \scriptsize{\{1217,} & \scriptsize{\{change,} & \scriptsize{\{-0.84,} & \scriptsize{\{geo,} & \scriptsize{\{-138,} & \scriptsize{\{1,57,}& \scriptsize{\{off\}} & \scriptsize{\{more,} & \scriptsize{\{off\}}\\
\scriptsize{simplex} & \scriptsize{7,9,10,11} & \scriptsize{on,lots,so\}} & \scriptsize{4\}} & \scriptsize{-3157,} & \scriptsize{1557,} & \scriptsize{exa\}} & \scriptsize{-0.62,} & \scriptsize{off\}} & \scriptsize{22,40,} & \scriptsize{251,270,}& \scriptsize{}& \scriptsize{off\}} & \scriptsize{}\\
\scriptsize{} & \scriptsize{15,20,25,} & \scriptsize{} & \scriptsize{} & \scriptsize{-3000,} & \scriptsize{1804,} & \scriptsize{} & \scriptsize{-0.35} & \scriptsize{} & \scriptsize{80,138\}} & \scriptsize{294\}}& \scriptsize{}& \scriptsize{} & \scriptsize{}\\
\scriptsize{} & \scriptsize{30,35,40,} & \scriptsize{} & \scriptsize{} & \scriptsize{-2395,} & \scriptsize{3384,} & \scriptsize{} & \scriptsize{0.62,0.66,} & \scriptsize{} & \scriptsize{} & \scriptsize{}& \scriptsize{}& \scriptsize{} & \scriptsize{}\\
\scriptsize{} & \scriptsize{50,60,80,} & \scriptsize{} & \scriptsize{} & \scriptsize{-2000,} & \scriptsize{4826\}} & \scriptsize{} & \scriptsize{0.84,0.91\}} & \scriptsize{} & \scriptsize{} & \scriptsize{}& \scriptsize{}& \scriptsize{} & \scriptsize{}\\
\scriptsize{} & \scriptsize{100\}} & \scriptsize{} & \scriptsize{} & \scriptsize{-1483,} & \scriptsize{} & \scriptsize{} & \scriptsize{} & \scriptsize{} & \scriptsize{} & \scriptsize{}& \scriptsize{}& \scriptsize{} & \scriptsize{}\\
\scriptsize{} & \scriptsize{} & \scriptsize{} & \scriptsize{} & \scriptsize{-1000,} & \scriptsize{} & \scriptsize{} & \scriptsize{} & \scriptsize{} & \scriptsize{} & \scriptsize{}& \scriptsize{}& \scriptsize{} & \scriptsize{}\\
\scriptsize{} & \scriptsize{} & \scriptsize{} & \scriptsize{} & \scriptsize{61\}} & \scriptsize{} & \scriptsize{} & \scriptsize{} & \scriptsize{} & \scriptsize{} & \scriptsize{}& \scriptsize{}& \scriptsize{} & \scriptsize{}\\
\hline
\scriptsize{default} & \scriptsize{-1} & \scriptsize{off} & \scriptsize{3} & \scriptsize{50} & \scriptsize{-1} & \scriptsize{auto} & \scriptsize{-0.5} & \scriptsize{auto} & \scriptsize{5} & \scriptsize{3} & \scriptsize{on} & \scriptsize{on} & \scriptsize{on}\\
\hline
\scriptsize{\textbf{$\mathcal{A}$}} & \scriptsize{\textbf{idiot}} & \scriptsize{\textbf{crash}} & \scriptsize{\textbf{dualize}} & \scriptsize{\textbf{pertv}} & \scriptsize{\textbf{sprint}} & \scriptsize{\textbf{dualp}} & \scriptsize{\textbf{psi}}& \scriptsize{\textbf{scal}} & \scriptsize{\textbf{passp}}
& \scriptsize{\textbf{subs}} & \scriptsize{\textbf{perturb}}& \scriptsize{\textbf{presolve}} & \scriptsize{\textbf{spars}}\\
\hline
\scriptsize{duals} & \scriptsize{} & \scriptsize{\{idiot1$\ldots$7,} & \scriptsize{\{1\}} & \scriptsize{\{-4900,} & \scriptsize{\{0,468,} & \scriptsize{\{pesteep,} & \scriptsize{\{-1.1,} & \scriptsize{\{geo,} & \scriptsize{\{-167,} & \scriptsize{\{37,}& \scriptsize{\{off\}} & \scriptsize{\{more,} & \scriptsize{\{off\}}\\
\scriptsize{simplex}& \scriptsize{} & \scriptsize{on,lots,so\}} & \scriptsize{} & \scriptsize{$\ldots$,820\}} & \scriptsize{620,} & \scriptsize{steep\}} & \scriptsize{$\ldots$,1.1\}} & \scriptsize{rows\}} & \scriptsize{-81,} & \scriptsize{40,41}& \scriptsize{}& \scriptsize{off\}} & \scriptsize{}\\
\scriptsize{} & \scriptsize{} & \scriptsize{} & \scriptsize{} & \scriptsize{} & \scriptsize{1612,} & \scriptsize{} & \scriptsize{} & \scriptsize{} & \scriptsize{-67,} & \scriptsize{297,}& \scriptsize{}& \scriptsize{} & \scriptsize{}\\
\scriptsize{} & \scriptsize{} & \scriptsize{} & \scriptsize{} & \scriptsize{} & \scriptsize{2228\}} & \scriptsize{} & \scriptsize{} & \scriptsize{} & \scriptsize{-33,} & \scriptsize{4354,}& \scriptsize{}& \scriptsize{} & \scriptsize{}\\
\scriptsize{} & \scriptsize{} & \scriptsize{} & \scriptsize{} & \scriptsize{} & \scriptsize{} & \scriptsize{} & \scriptsize{,} & \scriptsize{} & \scriptsize{0,36,67,} & \scriptsize{4392\}}& \scriptsize{}& \scriptsize{} & \scriptsize{}\\
\scriptsize{} & \scriptsize{} & \scriptsize{} & \scriptsize{} & \scriptsize{} & \scriptsize{} & \scriptsize{} & \scriptsize{} & \scriptsize{} & \scriptsize{93\}} & \scriptsize{}& \scriptsize{}& \scriptsize{} & \scriptsize{}\\
\hline
\scriptsize{default} & \scriptsize{} & \scriptsize{off} & \scriptsize{3} & \scriptsize{50} & \scriptsize{-1} & \scriptsize{auto} & \scriptsize{-0.5} & \scriptsize{auto} & \scriptsize{5} & \scriptsize{3} & \scriptsize{on} & \scriptsize{on} & \scriptsize{on}\\
\hline
\scriptsize{\textbf{$\mathcal{A}$}} & \scriptsize{\textbf{cholesky}} & \scriptsize{\textbf{gamma}} & \scriptsize{\textbf{dualize}} & \scriptsize{\textbf{pertv}} & \scriptsize{\textbf{sprint}} & \scriptsize{\textbf{dualp}} & \scriptsize{\textbf{psi}}& \scriptsize{\textbf{scal}} & \scriptsize{\textbf{passp}}
& \scriptsize{\textbf{subs}} & \scriptsize{\textbf{perturb}}& \scriptsize{\textbf{presolve}} & \scriptsize{\textbf{spars}}\\
\hline
\multirow{3}{*}{\scriptsize{barrier}}& \scriptsize{\{univ,} & \scriptsize{} &  & \scriptsize{\{-208,} & \scriptsize{} & \scriptsize{} & \scriptsize{} & \scriptsize{\{geo\}} & \scriptsize{\{83\}} & \scriptsize{\{132\}}& \scriptsize{} & \scriptsize{} & \scriptsize{}\\
 & \scriptsize{dense\}} & \scriptsize{} & \scriptsize{} & \scriptsize{-61,-50,} & \scriptsize{} & \scriptsize{} & \scriptsize{} & \scriptsize{} & \scriptsize{} & \scriptsize{}& \scriptsize{}& \scriptsize{} & \scriptsize{}\\
 & \scriptsize{} & \scriptsize{} & \scriptsize{} & \scriptsize{51,56,} & \scriptsize{} & \scriptsize{} & \scriptsize{} & \scriptsize{} & \scriptsize{} & \scriptsize{}& \scriptsize{}& \scriptsize{} & \scriptsize{}\\
  & \scriptsize{} & \scriptsize{} & \scriptsize{} & \scriptsize{61,102,} & \scriptsize{} & \scriptsize{} & \scriptsize{} & \scriptsize{} & \scriptsize{} & \scriptsize{}& \scriptsize{}& \scriptsize{} & \scriptsize{}\\
    & \scriptsize{} & \scriptsize{} & \scriptsize{} & \scriptsize{208\}} & \scriptsize{} & \scriptsize{} & \scriptsize{} & \scriptsize{} & \scriptsize{} & \scriptsize{}& \scriptsize{}& \scriptsize{} & \scriptsize{}\\
\hline
\scriptsize{default} & \scriptsize{native} & \scriptsize{off} & \scriptsize{3} & \scriptsize{50} & \scriptsize{} & \scriptsize{} & \scriptsize{} & \scriptsize{auto} & \scriptsize{5} & \scriptsize{3} & \scriptsize{on} & \scriptsize{on} & \scriptsize{on}\\
\hline
\end{tabular*}
\end{table}

\subsection{Experiments to evaluate scalability of the integer programming model}\label{firstExperiment}

\begin{figure}[H]
  \centerline{\includegraphics[width=10.5cm]{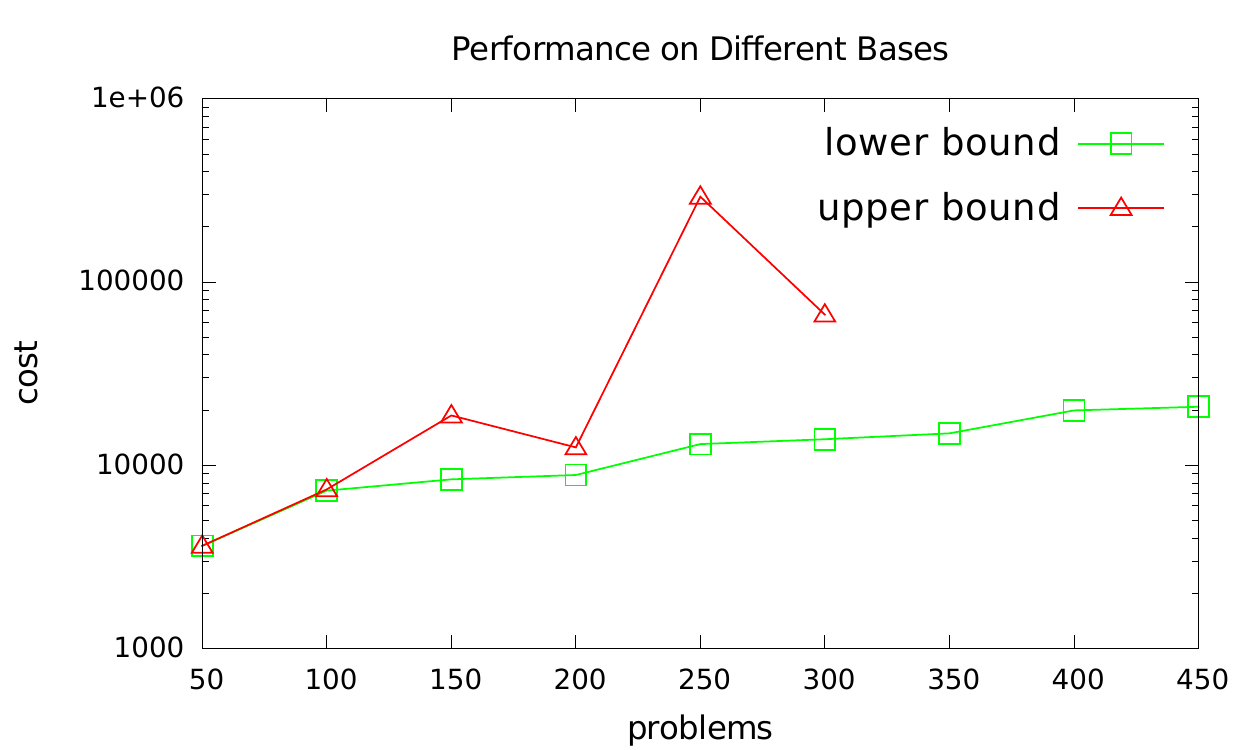}}
\caption{Performance of the integer programming model over sets of problem instances of different sizes and 50 algorithms.}
\label{scalabilityrandomproblems1}
\end{figure}

To evaluate the performance and the scalability of the proposed formulation in a standalone MIP solver, models for generating trees with different depths ($d=\{1,\ldots,3\}$) with datasets of different sizes built by randomly selecting subsets of results of the complete experimental results of the COIN-OR CBC solver were solved with the state-of-the-art CPLEX 12.9 MIP solver on a computer with 32GB of RAM and 6 Intel\textregistered i7-4960X cores. In this experiment, we measured the final gap reported by the solver between the best lower and upper bound at the end of execution with one hour time limit. These experiments considered generating trees with a minimum number of 10 instances per leaf node ($\tau = 10$) and penalty of ($\beta = 50$) for leaf nodes violating this constraint. Fig. \ref{scalabilityrandomproblems1} shows the performance of our integer programming model, considering bases with 50 algorithms and problem instances ranging from 50 to 500.

As it can be seen, optimal or near optimal decision trees were generated for models with up to 200 problem instances. For larger datasets the execution terminated with increasingly larger gaps for the produced bounds, at the point that for models with more than 300 instances a feasible solution was not found in the time limit. For the model with 500 problem instances not even the LP relaxation of the MIP model was computed in the time limit and no lower bound was available. Thus, for the complete dataset, experiments in the next subsections were performed only with the proposed VND-ASP heuristic.

\subsection{Experiments with the complete dataset}

To create optimized decision trees considering the entire experimental results dataset is quite challenging: there are more than half a million observations\footnote{534128 execution results produced by solving 1004 LP problems with 532 different algorithm configurations each}, far beyond the limits indicated in the previous section. Thus, only experiments with our mathematical programming heuristics were conducted for this dataset.

\begin{figure}[H]
\centerline{\includegraphics[scale=0.59]{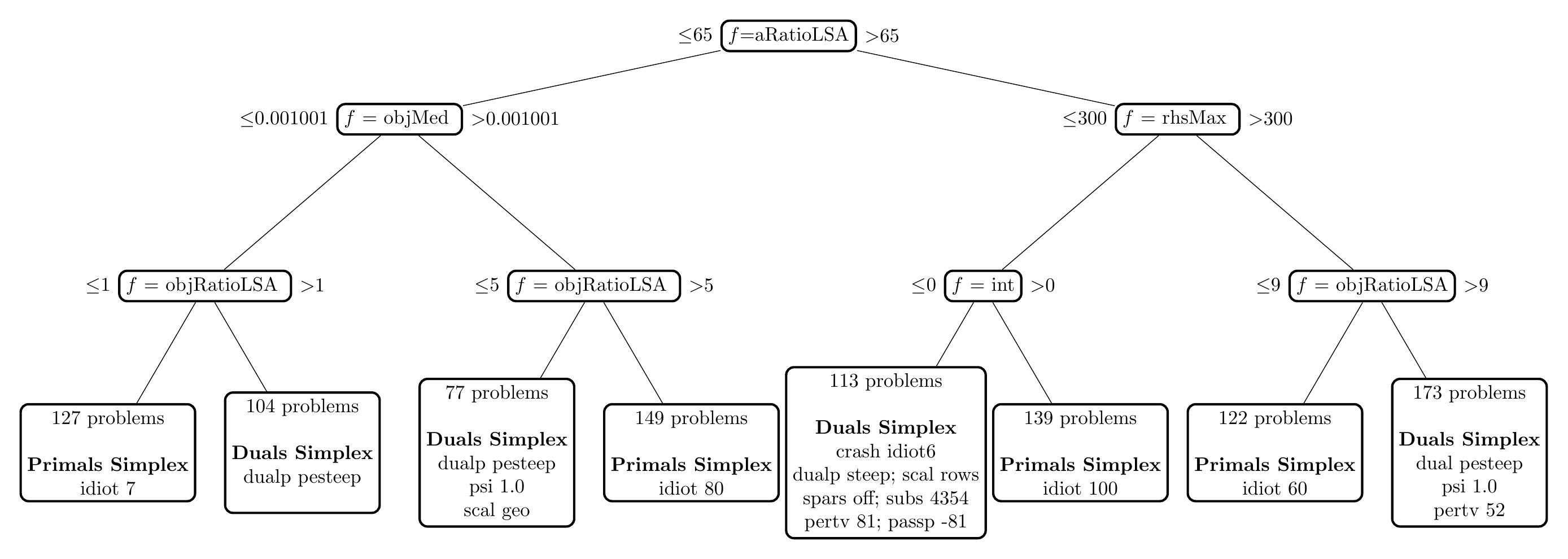}}
\caption{Decision tree with maximum depth = 3}
\label{tree}
\end{figure}

Fig. \ref{tree} presents the decision tree constructed with VNS-ASP using the following parameters: maximum tree depth $d=3$, total time limit $\mathcal{h} = 72000$, MIP search timeout $l=4000$, elite set size $m=20$, initial algorithms subsetsize $q=100$, $q'=20$, minimum number of instances per leaf node $\tau=50$ and penalty cost $\beta = 500$ . The estimated performance improvement with this decision tree is 61\%, a remarkable improvement. Please note, however, that this improvement does not reflects the expected performance improvement of this tree in unknown instances, which is the really important estimate. The estimated results of the  decision trees produced with our method on unknown instances is computed in the next section in 10-fold cross validation experiments.

An inspection in the contents of our decision tree shows that the range of the coefficients in the constraint matrix plays an important role for determining the best algorithm. The feature selected for the root node \textit{aRatioLSA} is computed as the ratio between the largest and the smallest absolute non-zero values in the constraint matrix. Each leaf node has a set of instances allocated to it, depending on the the decision on all parent nodes and a recommended algorithm, which is the algorithm with better results on these instances.  As an example, for the left-most branch of the tree, the best algorithm configuration select used the Primal simplex algorithm setting the ``idiot'' parameter  to value 7 considering 127 LP problems allocated to this node.

\subsection{Experiment using cross-validation on the complete base of problem instances}\label{secondExperiment}

To evaluate the predictive power of our method,  i.e. the expected performance on unknown instances, a 10-fold cross validation experiment was performed: a randomly shuffled complete dataset was  divided into 10 partitions and at each iteration 9 of the partitions were used to create the decision tree (training dataset) and the remaining partition used for evaluating the decision tree (test dataset). Each partition had 480928 examples (904 problem instances $\times$ 532 available algorithms), with the exception of the last four partitions that contained 480396 examples (903 problem instances $\times$ 532 available algorithms). The results of the cross-validation are given in Fig. 10. This figure shows the average performance degradation considering the ideal performance to solve the LP relaxation of all problem instances (the lower bound). Results of VND-ASP with maximum tree depth 4 (VND-ASP(D=5)) and 5 (VND-ASP(D=5)) are included. The remaining parameters of VND-ASP are the same described in the previous subsection. We also compare our results with the results produced with different configurations of the Random Forest (RF) algorithm implemented in Weka \citep{hall2009weka} (RF(T=1,$\ldots$,T=200, where T is the number of trees)). Default CBC settings (Default) and results obtained selecting a single best algorithm (SBA) are also included.

\begin{figure}[H]
\begin{center}
\includegraphics[width=16cm]{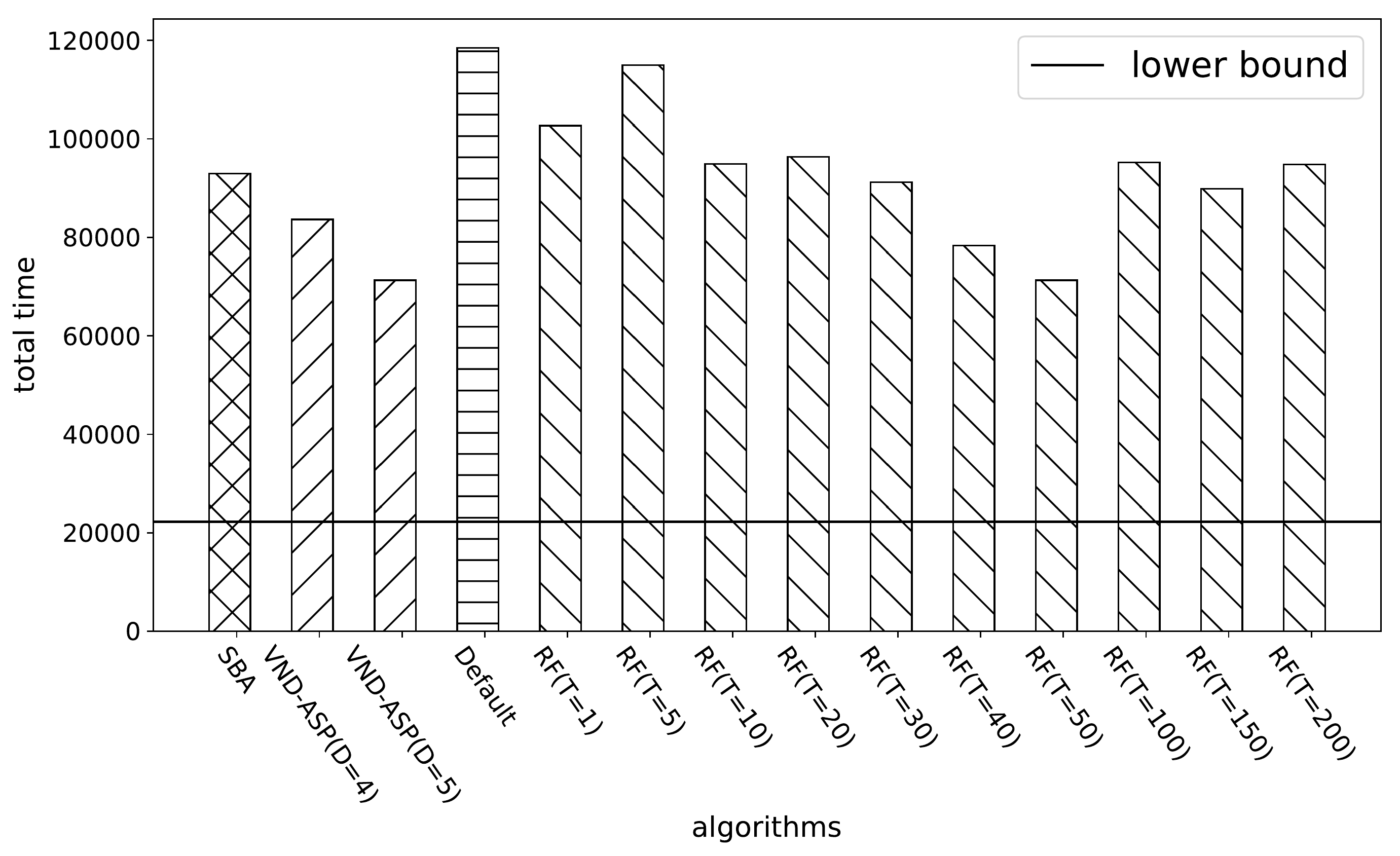}
\label{fig:crossresults} \caption{Cross-validation results for all partitions}
\end{center}
\end{figure}

As can be seen, our results indicate performance gains of 40\% compared against default settings. Moreover, they are noticeably better than those obtained when selecting only the single best algorithm (22\%). VND-ASP results are also mostly superior to the ones obtained with RF, with the exception of RF(T=50), where equivalent results were obtained.  We believe that this is a very positive result, since the result produced by our algorithm (a single tree) is more easy to interpret than those produced by RF. More importantly, algorithms can be recommended much faster (in constant time) with our approach since the processing cost does not depends on the number of available algorithms as in RF, where a series of regression problems must be solved in order to recommend an algorithm for each new instance.

\section{Discussion and closing remarks}\label{sec:Conclusions}

This paper introduced a new mathematical programming formulation to solve the Algorithm Selection Problem (ASP). This formulation produces globally optimal decision trees with limited depth. The main advantage of this approach is that, despite the construction of the tree itself potentially being computationally expensive, once the tree has been constructed, algorithm recommendations can be made in constant time. A dataset containing the experimental results of many linear programming solver configurations of the COIN-OR Branch-\&-Cut linear programming solver (CLP) was built solving a comprehensive set of instances from various applications. This initial batch of experiments itself already revealed improved parameter settings for the LP solver, including the discovery of a new algorithm configuration which was 22\% faster than default CLP settings.

Scalability tests were performed to check how large subsets could become before it was no longer possible to generate provably optimal decision trees with a state of the art standalone MIP solver. Given that, at a certain point, the resulting MIP model becomes too difficult to optimize exactly, a mathematical programming-based VND local search heuristic was also proposed to handle larger datasets.

To evaluate the predictive power of our method, a 10-fold cross validation experiment was conducted. The results were very promising: executions with the recommended parameter settings were 40\% faster than CLP default settings, almost doubling the improvement that could be obtained using a single best parameter setting. Our results are comparable with those obtained after tuning the Random Forest algorithm, with the added advantage that the predictive model produced by our method (a single tree) is easily interpretable and, more importantly, the cost of recommending an algorithm is not dependent upon the number of available algorithms.

Future directions include evaluating stronger alternative integer programming formulations for this problem given that, as the scalability test showed, there is still a significant gap between the lower and upper bounds produced for the larger datasets. The positive results for the ASP are also a good indicator that the application of our methodology to classification and regression problems represents a promising future research path.

\section*{Acknowledgements}
The authors would like to thank the Brazilian agencies CNPq and FAPEMIG
for the financial support. The authors acknowledge the National Laboratory
for Scientific Computing (LNCC/MCTI, Brazil) for providing HPC resources
of the SDumont supercomputer, which have contributed to the research
results reported within this paper. URL: http://sdumont.lncc.br. The research was partially supported by Data-driven logistics (FWO-S007318N). Editorial consultation provided by Luke Connolly (KU Leuven).


\nocite{*}

\bibliographystyle{itor}
\bibliography{opt-dtree-asp}

\begin{thebibliography}{62}
\expandafter\ifx\csname natexlab\endcsname\relax\def\natexlab#1{#1}\fi
\expandafter\ifx\csname url\endcsname\relax
  \def\url#1{\texttt{#1}}\fi
\expandafter\ifx\csname urlprefix\endcsname\relax\def\urlprefix{URL }\fi
\providecommand{\eprint}[2][]{\url{#2}}
\bibitem[{Achard and De~Schutter(2006)}]{achard}
Achard, P., De~Schutter, E., 2006.
\newblock Complex {P}arameter {L}andscape for a {C}omplex {N}euron {M}odel.
\newblock \textit{PLOS Computational Biology} 2, 7, 1--11.
\bibitem[{Applegate et~al.(2006)Applegate, Bixby, Chvatal and
  Cook}]{applegate2006traveling}
Applegate, D.L., Bixby, R.E., Chvatal, V., Cook, W.J., 2006.
\newblock \textit{The traveling salesman problem: a computational study}.
\newblock Princeton university press.
\bibitem[{Atamt{\"u}rk and Savelsbergh(2005)}]{atamturk2005integer}
Atamt{\"u}rk, A., Savelsbergh, M.W., 2005.
\newblock Integer-programming software systems.
\newblock \textit{Annals of operations research} 140, 1, 67--124.
\bibitem[{Atkeson et~al.(1997)Atkeson, Moore and
  Schaal}]{Atkeson97locallyweighted}
Atkeson, C.G., Moore, A.W., Schaal, S., 1997.
\newblock Locally {W}eighted {L}earning.
\newblock \textit{Artificial Intelligence Review} 11, 1-5, 11--73.
\bibitem[{Audet and Orban(2006)}]{audet}
Audet, C., Orban, D., 2006.
\newblock Finding {O}ptimal {A}lgorithmic {P}arameters {U}sing
  {D}erivative-{F}ree {O}ptimization.
\newblock \textit{SIAM Journal on Optimization} 17, 3, 642--664.
\bibitem[{Battistutta et~al.(2017)Battistutta, Schaerf and
  Urli}]{battistutta2017feature}
Battistutta, M., Schaerf, A., Urli, T., 2017.
\newblock Feature-based tuning of single-stage simulated annealing for
  examination timetabling.
\newblock \textit{Annals of Operations Research} 252, 2, 239--254.
\bibitem[{Baz et~al.(2007)Baz, Hunsaker, Brooks and Gosavi}]{baz2007automated}
Baz, M., Hunsaker, B., Brooks, J.P., Gosavi, A., 2007.
\newblock Automated {T}uning of {O}ptimization {S}oftware {P}arameters.
\newblock Technical report, Technical Report TR2007-7, University of
  Pittsburgh, Department of Industrial Engineering.
\bibitem[{Bellio et~al.(2016)Bellio, Ceschia, Gaspero, Schaerf and
  Urli}]{Bellio201683}
Bellio, R., Ceschia, S., Gaspero, L.D., Schaerf, A., Urli, T., 2016.
\newblock Feature-{B}ased {T}uning of {S}imulated {A}nnealing applied to the
  {C}urriculum-{B}ased {C}ourse {T}imetabling {P}roblem.
\newblock \textit{Computers \& Operations Research} 65, 83 -- 92.
\bibitem[{Bertsimas and Dunn(2017)}]{Bertsimas:2017:OCT:3123655.3123731}
Bertsimas, D., Dunn, J., 2017.
\newblock Optimal {C}lassification {T}rees.
\newblock \textit{Machine Learning} 106, 7, 1039--1082.
\bibitem[{Bixby et~al.(2004)Bixby, Fenelon, Gu, Rothberg and
  Wunderling}]{bixbyprogress2004}
Bixby, R.E., Fenelon, M., Gu, Z., Rothberg, E., Wunderling, R., 2004.
\newblock Mixed {I}nteger {P}rogramming: {A} {P}rogress {R}eport.
\newblock In Gr{\"o}tschel, M. (ed.), \textit{The Sharpest Cut: The Impact of
  Manfred Padberg and His Work}.
\newblock SIAM, chapter~18, pp. 309--324.
\bibitem[{Bolme et~al.(2011)Bolme, Beveridge, Draper, Phillips and
  Lui}]{Bolme2011}
Bolme, D.S., Beveridge, J.R., Draper, B.A., Phillips, P.J., Lui, Y.M., 2011.
\newblock Automatically {S}earching for {O}ptimal {P}arameter {S}ettings
  {Using} a {G}enetic {A}lgorithm.
\newblock In Crowley, J.L., Draper, B.A. and Thonnat, M. (eds),
  \textit{Computer Vision Systems}, Springer Berlin Heidelberg, Berlin,
  Heidelberg, pp. 213--222.
\bibitem[{Breiman(2001)}]{rf}
Breiman, L., 2001.
\newblock Random {F}orests.
\newblock \textit{Machine Learning} 45, 1, 5--32.
\bibitem[{Breiman et~al.(1984)Breiman, Friedman, Olshen and Stone}]{breiman}
Breiman, L., Friedman, J.H., Olshen, R.A., Stone, C.J., 1984.
\newblock \textit{{Classification and Regression Trees}}.
\newblock Statistics/Probability Series.
\newblock Wadsworth Publishing Company, Belmont, California, U.S.A.
\bibitem[{Eiben et~al.(1999)Eiben, Hinterding and
  Michalewicz}]{Eiben:1999:PCE:2221345.2221455}
Eiben, A.E., Hinterding, R., Michalewicz, Z., 1999.
\newblock Parameter {C}ontrol in {E}volutionary {A}lgorithms.
\newblock \textit{Transactions on Evolutionary Computation} 3, 2, 124--141.
\bibitem[{Fischetti and Fischetti(2016)}]{fischetti2016matheuristics}
Fischetti, M., Fischetti, M., 2016.
\newblock Matheuristics.
\newblock \textit{Handbook of Heuristics}
\newblock pp. 1--33.
\bibitem[{Fonseca et~al.(2017)Fonseca, Santos, Carrano and
  Stidsen}]{Fonseca2017}
Fonseca, G.H., Santos, H.G., Carrano, E.G., Stidsen, T.J., 2017.
\newblock Integer programming techniques for educational timetabling.
\newblock \textit{European Journal of Operational Research} 262, 28--39.
\bibitem[{Forrest and Lougee-Heimer(2005)}]{forrest2005cbc}
Forrest, J., Lougee-Heimer, R., 2005.
\newblock C{B}{C} {U}ser {G}uide.
\newblock In \textit{Emerging Theory, Methods, and Applications}.
\newblock INFORMS, pp. 257--277.
\bibitem[{Gamrath et~al.(2015)Gamrath, Koch, Martin, Miltenberger and
  Weninger}]{Gamrath2013}
Gamrath, G., Koch, T., Martin, A., Miltenberger, M., Weninger, D., 2015.
\newblock {Progress in Presolving for Mixed Integer Programming}.
\newblock \textit{Mathematical Programming Computation} 7, 367--398.
\bibitem[{Garey and Johnson(1979)}]{garey2002computers}
Garey, M.R., Johnson, D.S., 1979.
\newblock \textit{Computers and {I}ntractability; {A} {G}uide to the {T}heory
  of {N}P-{C}ompleteness}.
\newblock W. H. Freeman \& Co., New York, NY, USA.
\bibitem[{Gearhart et~al.(2013)Gearhart, Adair, Detry, Durfee, Jones and
  Martin}]{Gearhart13}
Gearhart, J.L., Adair, K.L., Detry, R.J., Durfee, J.D., Jones, K.A., Martin,
  N., 2013.
\newblock Comparison of Open-Source Linear Programming Solvers.
\newblock Technical report, Sandia National Laboratories.
\bibitem[{Haas et~al.(2005)Haas, Peysakhov and Mancoridis}]{haas}
Haas, J., Peysakhov, M., Mancoridis, S., 2005.
\newblock G{A}-{B}ased {P}arameter {T}uning for {M}ulti-{A}gent {S}ystems.
\newblock In \textit{Proceedings of the 7th Annual Conference on Genetic and
  Evolutionary Computation}, ACM, New York, NY, USA, pp. 1085--1086.
\bibitem[{Hall et~al.(2009)Hall, Frank, Holmes, Pfahringer, Reutemann and
  Witten}]{hall2009weka}
Hall, M., Frank, E., Holmes, G., Pfahringer, B., Reutemann, P., Witten, I.H.,
  2009.
\newblock The weka data mining software: an update.
\newblock \textit{ACM SIGKDD explorations newsletter} 11, 1, 10--18.
\bibitem[{Hutter et~al.(2014a)Hutter, St{\"{u}}tzle, Leyton{-}Brown and
  Hoos}]{DBLP:journals/corr/HutterSLH14}
Hutter, F., St{\"{u}}tzle, T., Leyton{-}Brown, K., Hoos, H.H., 2014a.
\newblock Paramils: An automatic algorithm configuration framework.
\newblock \textit{CoRR} abs/1401.3492.
\newblock \eprint{1401.3492}.
\bibitem[{Hutter et~al.(2014b)Hutter, Xu, H. and Leyton-Brown}]{hutter201479}
Hutter, F., Xu, L., H., H.H., Leyton-Brown, K., 2014b.
\newblock Algorithm runtime prediction: Methods \& evaluation.
\newblock \textit{Artificial Intelligence} 206, 79 -- 111.
\bibitem[{Hyafil and Rivest(1976)}]{HYAFIL197615}
Hyafil, L., Rivest, R.L., 1976.
\newblock Constructing {O}ptimal {B}inary {D}ecision {T}rees is
  {N}{P}-{C}omplete.
\newblock \textit{Information Processing Letters} 5, 1, 15 -- 17.
\bibitem[{Johnson et~al.(2000)Johnson, Nemhauser and Savelsbergh}]{Johnson2000}
Johnson, E., Nemhauser, G., Savelsbergh, W., 2000.
\newblock {Progress in Linear Programming-Based Algorithms for Integer
  Programming: An Exposition}.
\newblock \textit{INFORMS Journal on Computing} 12.
\bibitem[{Kadioglu et~al.(2010)Kadioglu, Malitsky, Sellmann and
  Tierney}]{Kadioglu:2010:IIA:1860967.1861114}
Kadioglu, S., Malitsky, Y., Sellmann, M., Tierney, K., 2010.
\newblock Isac --instance-specific algorithm configuration.
\newblock In \textit{Proceedings of the 2010 Conference on ECAI 2010: 19th
  European Conference on Artificial Intelligence}, IOS Press, Amsterdam, The
  Netherlands, The Netherlands, pp. 751--756.
\bibitem[{Kass(1980)}]{kass}
Kass, G.V., 1980.
\newblock An {E}xploratory {T}echnique for {I}nvestigating {L}arge {Q}uantities
  of {C}ategorical {D}ata.
\newblock \textit{Journal of the Royal Statistical Society. Series C (Applied
  Statistics)} 29, 2, 119--127.
\bibitem[{Koch et~al.(2011)Koch, Achterberg, Andersen, Bastert, Berthold,
  Bixby, Danna, Gamrath, Gleixner, Heinz et~al.}]{koch2011miplib}
Koch, T., Achterberg, T., Andersen, E., Bastert, O., Berthold, T., Bixby, R.E.,
  Danna, E., Gamrath, G., Gleixner, A.M., Heinz, S., et~al., 2011.
\newblock {MIPLIB} 2010.
\newblock \textit{Mathematical Programming Computation} 3, 2, 103.
\bibitem[{Kohavi and John(1995)}]{kohavi}
Kohavi, R., John, G.H., 1995.
\newblock Automatic {P}arameter {S}election by {M}inimizing {E}stimated
  {E}rror.
\newblock In \textit{In Proceedings of the Twelfth International Conference on
  Machine Learning}, Morgan Kaufmann, pp. 304--312.
\bibitem[{Land and Doig(1960)}]{doig1960}
Land, A.H., Doig, A.G., 1960.
\newblock An {A}utomatic {M}ethod for {S}olving {D}iscrete {P}rogramming
  {P}roblems.
\newblock \textit{Econometrica} 28, 3, 497--520.
\bibitem[{Leyton-Brown et~al.(2003a)Leyton-Brown, Nudelman, Andrew, McFadden
  and Shoham}]{10.1007/978-3-540-45193-8_75}
Leyton-Brown, K., Nudelman, E., Andrew, G., McFadden, J., Shoham, Y., 2003a.
\newblock Boosting as a metaphor for algorithm design.
\newblock In Rossi, F. (ed.), \textit{Principles and Practice of Constraint
  Programming -- CP 2003}, Springer Berlin Heidelberg, Berlin, Heidelberg, pp.
  899--903.
\bibitem[{Leyton-Brown et~al.(2003b)Leyton-Brown, Nudelman, Andrew, McFadden
  and Shoham}]{leyton2003portfolio}
Leyton-Brown, K., Nudelman, E., Andrew, G., McFadden, J., Shoham, Y., 2003b.
\newblock A portfolio approach to algorithm selection.
\newblock In \textit{IJCAI}, Vol.~3, pp. 1542--1543.
\bibitem[{Loh and Shih(1997)}]{shih}
Loh, W.Y., Shih, Y.S., 1997.
\newblock Split selection methods for classification trees.
\newblock \textit{Statistica Sinica} 7, 4, 815--840.
\bibitem[{L{\'o}pez-Ib{\'a}{\~n}ez et~al.(2016)L{\'o}pez-Ib{\'a}{\~n}ez,
  Dubois-Lacoste, {P{\'e}rez C{\'a}ceres}, St{\"u}tzle and Birattari}]{irace}
L{\'o}pez-Ib{\'a}{\~n}ez, M., Dubois-Lacoste, J., {P{\'e}rez C{\'a}ceres}, L.,
  St{\"u}tzle, T., Birattari, M., 2016.
\newblock The irace {P}ackage: {I}terated {R}acing for {A}utomatic {A}lgorithm
  {C}onfiguration.
\newblock \textit{Operations Research Perspectives} 3, 43--58.
\bibitem[{L{\'o}pez-Ib{\'a}{\~{n}}ez and St{\"u}tzle(2014)}]{ibanez}
L{\'o}pez-Ib{\'a}{\~{n}}ez, M., St{\"u}tzle, T., 2014.
\newblock Automatically {I}mproving the {A}nytime {B}ehaviour of {O}ptimisation
  {A}lgorithms.
\newblock \textit{European Journal of Operational Research} 235, 3, 569 -- 582.
\bibitem[{Lougee-Heimer(2003)}]{Lougee-Heimer2003}
Lougee-Heimer, R., 2003.
\newblock {The Common Optimization Interface for Operations Research: Promoting
  Open-Source Software in the Operations Research Community}.
\newblock \textit{IBM Journal of Research and Development} 47, 1, 57--66.
\bibitem[{Mascia et~al.(2014)Mascia, Pellegrini, Birattari and
  Stützle}]{mascia2014an}
Mascia, F., Pellegrini, P., Birattari, M., Stützle, T., 2014.
\newblock An {A}nalysis of {P}arameter {A}daptation in {R}eactive {T}abu
  {S}earch.
\newblock \textit{International Transactions in Operational Research} 21, 1,
  127--152.
\bibitem[{Menickelly et~al.(2016)Menickelly, G{\"{u}}nl{\"{u}}k, Kalagnanam and
  Scheinberg}]{DBLP:journals/corr/MenickellyGKS16}
Menickelly, M., G{\"{u}}nl{\"{u}}k, O., Kalagnanam, J., Scheinberg, K., 2016.
\newblock Optimal {G}eneralized {D}ecision {T}rees via {I}nteger {P}rogramming.
\newblock \textit{CoRR} abs/1612.03225.
\bibitem[{Misir and Sebag(2017)}]{MISIR2017291}
Misir, M., Sebag, M., 2017.
\newblock Alors: An algorithm recommender system.
\newblock \textit{Artificial Intelligence} 244, 291 -- 314.
\bibitem[{Mittelmann(2018)}]{hansmtbenchlp}
Mittelmann, H., 2018.
\newblock Benchmark of simplex {LP} solvers.
\newblock \url{http://plato.asu.edu/ftp/lpsimp.html}.
\newblock Accessed: 2018-10-03.
\bibitem[{Mladenovi\'{c} and Hansen(1997)}]{vndref1}
Mladenovi\'{c}, N., Hansen, P., 1997.
\newblock Variable {N}eighborhood {S}earch.
\newblock \textit{Computers and Operations Research} 24, 11, 1097--1100.
\bibitem[{Pochet and Wolsey(2006)}]{Pochet:2006:PPM:1202598}
Pochet, Y., Wolsey, L.A., 2006.
\newblock \textit{Production Planning by Mixed Integer Programming (Springer
  Series in Operations Research and Financial Engineering)}.
\newblock Springer-Verlag New York, Inc., Secaucus, NJ, USA.
\bibitem[{Polyakovskiy et~al.(2014)Polyakovskiy, Bonyadi, Wagner, Michalewicz
  and Neumann}]{polyakovskiy2014comprehensive}
Polyakovskiy, S., Bonyadi, M.R., Wagner, M., Michalewicz, Z., Neumann, F.,
  2014.
\newblock A comprehensive benchmark set and heuristics for the traveling thief
  problem.
\newblock In \textit{Proceedings of the 2014 Annual Conference on Genetic and
  Evolutionary Computation}, \bibinfo{organization}{ACM}, pp. 477--484.
\bibitem[{Quinlan(1986)}]{Quinlan:1986:IDT:637962.637969}
Quinlan, J.R., 1986.
\newblock Induction of {D}ecision {T}rees.
\newblock \textit{Machine Learning} 1, 1, 81--106.
\bibitem[{Quinlan(1993)}]{Quinlan:1993:CPM:152181}
Quinlan, J.R., 1993.
\newblock \textit{C4.5: Programs for Machine Learning}.
\newblock Morgan Kaufmann Publishers Inc., San Francisco, CA, USA.
\bibitem[{Resende and Ribeiro(2014)}]{graspref}
Resende, M., Ribeiro, C., 2014.
\newblock GRASP: Greedy randomized adaptive search procedures, Springer US.
\newblock pp. 287--312.
\bibitem[{Rice(1976)}]{RICE197665}
Rice, J.R., 1976.
\newblock The algorithm selection problem.
\newblock  \textit{Advances in Computers}. Vol.~15.
\newblock Elsevier, pp. 65 -- 118.
\bibitem[{Santos et~al.(2016)Santos, Toffolo, Gomes and Ribas}]{Santos2016}
Santos, H.G., Toffolo, T.A., Gomes, R.A., Ribas, S., 2016.
\newblock Integer programming techniques for the nurse rostering problem.
\newblock \textit{Annals of Operations Research} 239, 225--251.
\bibitem[{Silva and Santos(2017)}]{esilva2017}
Silva, C., Santos, H., 2017.
\newblock Drawing graphs with mathematical programming and variable
  neighborhood search.
\newblock \textit{Electronic Notes in Discrete Mathematics} 58, 207 -- 214.
\bibitem[{Song and Lu(2015)}]{yan}
Song, Y.Y., Lu, Y., 2015.
\newblock Decision tree methods: applications for classification and
  prediction,
\newblock 27, 2, 130--135.
\bibitem[{Souza et~al.(2010)Souza, Coelho, Ribas, Santos and
  Merschmann}]{SOUZA20101041}
Souza, M., Coelho, I., Ribas, S., Santos, H., Merschmann, L., 2010.
\newblock A hybrid heuristic algorithm for the open-pit-mining operational
  planning problem.
\newblock \textit{European Journal of Operational Research} 207, 2, 1041 --
  1051.
\bibitem[{Tange(2011)}]{tange2011gnu}
Tange, O., 2011.
\newblock Gnu parallel-the command-line power tool.
\newblock \textit{The USENIX Magazine} 36, 1, 42--47.
\bibitem[{Tsoumakas and Katakis(2007)}]{tsoumakas2007multi}
Tsoumakas, G., Katakis, I., 2007.
\newblock Multi-label classification: An overview.
\newblock \textit{International Journal of Data Warehousing and Mining (IJDWM)}
  3, 3, 1--13.
\bibitem[{Vapnik(1995)}]{Vapnik1995}
Vapnik, V.N., 1995.
\newblock \textit{The Nature of Statistical Learning Theory}.
\newblock Springer-Verlag New York, Inc., New York, NY, USA.
\bibitem[{Vilas~Boas et~al.(2017)Vilas~Boas, Santos, Martins and
  Merschmann}]{Boas2017715}
Vilas~Boas, M.G., Santos, H.G., Martins, R.S.O., Merschmann, L.H.C., 2017.
\newblock Data {M}ining {A}pproach for {F}eature {B}ased {P}arameter {T}unning
  for {M}ixed-{I}nteger {P}rogramming {S}olvers.
\newblock \textit{Procedia Computer Science} 108, 715 -- 724.
\bibitem[{Witten et~al.(2011)Witten, Frank and Hall}]{Witten20113}
Witten, I.H., Frank, E., Hall, M.A., 2011.
\newblock \textit{Data {M}ining: {P}ractical {M}achine {L}earning {T}ools and
  {T}echniques}  (3rd edn.).
\newblock Morgan Kaufmann Publishers Inc., San Francisco, CA, USA.
\bibitem[{Wolsey(2007)}]{wolsey08}
Wolsey, L.A., 2007.
\newblock Mixed Integer Programming, John Wiley \& Sons, Inc.
\bibitem[{Xu et~al.(2010)Xu, Hoos and Leyton-Brown}]{AAAI101929}
Xu, L., Hoos, H., Leyton-Brown, K., 2010.
\newblock Hydra: Automatically configuring algorithms for portfolio-based
  selection.
\bibitem[{Xu et~al.(2008)Xu, Hutter, Hoos and Leyton-Brown}]{xu2008satzilla}
Xu, L., Hutter, F., Hoos, H.H., Leyton-Brown, K., 2008.
\newblock Satzilla: portfolio-based algorithm selection for sat.
\newblock \textit{Journal of artificial intelligence research} 32, 565--606.
\bibitem[{Zhang et~al.(2018)Zhang, Wang, Xu, Chow and
  Wu}]{zhang2018tree2vector}
Zhang, H., Wang, S., Xu, X., Chow, T.W., Wu, Q.J., 2018.
\newblock Tree2vector: learning a vectorial representation for tree-structured
  data.
\newblock \textit{IEEE transactions on neural networks and learning systems}
\newblock , 99, 1--15.
\bibitem[{Zhu(2007)}]{zhu}
Zhu, Y., 2007.
\newblock Mixed-{I}nteger {L}inear {P}rogramming {A}lgorithm for a
  {C}omputational {P}rotein {D}esign {P}roblem.
\newblock \textit{Industrial \& Engineering Chemistry Research} 46, 3,
  839--845.

\end{thebibliography}

\end{document}